\documentclass[a4paper,fleqn]{cas-dc}

\usepackage[numbers,sort&compress]{natbib}
\usepackage{xspace}
\usepackage{multirow}
\usepackage{graphicx}
\usepackage{inconsolata}
\usepackage{booktabs}
\usepackage{subfig}
\usepackage{amsfonts}
\usepackage[ruled, lined, linesnumbered, commentsnumbered, longend]{algorithm2e}
\usepackage[normalem]{ulem}
\usepackage[title]{appendix}
\usepackage[normalem]{ulem} 
\usepackage{amsmath}
\usepackage[flushleft]{threeparttable} 

\newcommand{\secref}[1]{\S\ref{#1}}

\newcommand{\SQUADTR}{\mbox{\texttt{SQuAD-TR}}\xspace}
\newcommand{\SQUADTRTRAIN}{\mbox{\texttt{SQuAD-TR-TRAIN}}\xspace}

\newcommand{\SQUADTRTRAINCOLBERTYYYY}{\mbox{\texttt{SQuAD-TR-TRAIN (ColBERT-QA,YYYY)}}\xspace}
\newcommand{\SQUADTRTRAINCOLBERTYYYYOLD}{\mbox{\texttt{SQuAD-TR-TRAIN (ColBERT-QA,2021)}}\xspace}
\newcommand{\SQUADTRTRAINCOLBERTYYYYNEW}{\mbox{\texttt{SQuAD-TR-TRAIN (ColBERT-QA,2023)}}\xspace}

\newcommand{\SQUADTRTRAINBMYYYY}{\mbox{\texttt{SQuAD-TR-TRAIN (BM25,YYYY)}}\xspace}
\newcommand{\SQUADTRTRAINBMYYYYOLD}{\mbox{\texttt{SQuAD-TR-TRAIN (BM25,2021)}}\xspace}
\newcommand{\SQUADTRTRAINBMYYYYNEW}{\mbox{\texttt{SQuAD-TR-TRAIN (BM25,2023)}}\xspace}

\newcommand{\SQUADTRTRAINDPRYYYY}{\mbox{\texttt{SQuAD-TR-TRAIN (DPR,YYYY)}}\xspace}

\newcommand{\SQUADTRDEV}{\mbox{\texttt{SQuAD-TR-DEV}}\xspace}

\newcommand{\XQUAD}{\texttt{XQuAD}}
\newcommand{\XQUADTR}{\mbox{\texttt{XQuAD-TR}}\xspace}

\newcommand{\XQUADTRCOLBERTYYYYOLD}{\mbox{\texttt{XQuAD-TR (ColBERT-QA,2021)}}\xspace}
\newcommand{\XQUADTRCOLBERTYYYYNEW}{\mbox{\texttt{XQuAD-TR (ColBERT-QA,2023)}}\xspace}

\newcommand{\XQUADTRBMYYYYOLD}{\mbox{\texttt{XQuAD-TR (BM25,2021)}}\xspace}
\newcommand{\XQUADTRBMYYYYNEW}{\mbox{\texttt{XQuAD-TR (BM25,2023)}}\xspace}

\newcommand{\SQUADEN}{\mbox{\texttt{SQuAD-EN}}\xspace}
\newcommand{\SQUADENTRAIN}{\mbox{\texttt{SQuAD-EN-TRAIN}}\xspace}
\newcommand{\SQUADENDEV}{\mbox{\texttt{SQuAD-EN-DEV}}\xspace}
\newcolumntype{P}[1]{>{\raggedright\arraybackslash}p{#1}}
\newcommand{\COLBERT}{\mbox{ColBERT}}
\newcommand{\COLBERTQA}{\mbox{\COLBERT-QA}\xspace}

\newcommand{\XLMROBERTA}{\mbox{XLM-RoBERTa}\xspace}
\newcommand{\RCONE}{\mbox{C@1}\xspace}
\newcommand{\RCFIVE}{\mbox{C@5}\xspace}
\newcommand{\RCTWENTY}{\mbox{C@20}\xspace}
\newcommand{\WIKIEN}{\mbox{\texttt{Wiki-EN-2018}}\xspace}
\newcommand{\WIKITR}{\mbox{\texttt{Wiki-TR}}\xspace}
\newcommand{\WIKITROLD}{\mbox{\texttt{Wiki-TR-2021}}\xspace}
\newcommand{\WIKITRNEW}{\mbox{\texttt{Wiki-TR-2023}}\xspace}

\newcommand{\SQUADTRURLPROD}{\mbox{\url{https://github.com/boun-tabi/SQuAD-TR}}\xspace}
\newcommand{\REVISIONCOLOR}{\color{black}}

\newcommand{\SQUADTRURL}{{\REVISIONCOLOR\underline{\SQUADTRURLPROD}}}

\newcommand{\ANSWERCOLOR}{\color{purple}}
\newcommand{\ANSWERCOLORINREVISION}{\color{purple}}
\newcommand{\CUSTOMVSPACE}{\vspace{3.5pt}}

\newcommand{\DPRBASED}{\mbox{DPR-based}}
\newcommand{\COLBERTQABASED}{\mbox{\COLBERTQA-based}}

\NewDocumentCommand \printassistiveaitechdisclosure {} { 
\section*{Declaration of AI and AI-assisted technologies in the writing process}
During the preparation of this work the corresponding author used the ChatGPT service for proofreading, spell checking, and grammar correction in specific sections of the manuscript. After using this service, the author carefully reviewed and edited the content as needed, and all the authors take full responsibility for the content of the publication.
}

\def\tsc#1{\csdef{#1}{\textsc{\lowercase{#1}}\xspace}}
\tsc{WGM}
\tsc{QE}
\tsc{EP}
\tsc{PMS}
\tsc{BEC}
\tsc{DE}

\begin{document}
    \let\WriteBookmarks\relax
    \def\floatpagepagefraction{1}
    \def\textpagefraction{.001}

    \title [mode = title]{Building Efficient and Effective OpenQA Systems for Low-Resource Languages}
    
    \shorttitle{Efficient and Effective OpenQA}
    
    \shortauthors{Budur et~al.}
    
    \author[1,2]{Emrah Budur}[type=editor]
    \cormark[1] 
    \ead{emrah.budur@std.bogazici.edu.tr}
    \fnmark[1]
    \credit{Methodology, Software, Data Curation, Formal Analysis, Investigation, Validation, Visualization, Writing -- original draft, Funding acquisition}
    
    \author[3]{R{\i}za \"{O}z\c{c}elik}[]
    \ead{r.ozcelik@tue.nl}
    \credit{Data Curation, Formal analysis, Writing -- review \& editing, Funding acquisition}
    
    \author[4]{Dilara Soylu}[]
    \ead{soylu@stanford.edu}
    \credit{Software, Writing -- original draft}
    
    \author[4]{Omar Khattab}[]
    \ead{okhattab@stanford.edu}
    \credit{Conceptualization, Methodology, Software, Validation, Writing -- review \& editing}
    
    \author[1]{Tunga G\"{u}ng\"{o}r}[]
    \ead{gungort@bogazici.edu.tr}
    \credit{Conceptualization, Methodology, Formal analysis, Writing -- review \& editing, Supervision}
    
    \author[4]{Christopher Potts}[]
    \ead{cgpotts@stanford.edu}
    \credit{Conceptualization, Methodology, Formal Analysis, Writing -- review \& editing, Supervision}
    
    \cortext[cor1]{Corresponding author}
    
    \fntext[fn1]{Work done outside of Amazon.}
      
    \affiliation[1]{organization={Bo\u{g}azi\c{c}i University},
        addressline={Bebek}, 
        city={Istanbul},
        postcode={34342},
        country={Turkey}}
        
    \affiliation[2]{organization={Amazon},
        city={Toronto},
        state={ON},
        country={Canada}}

    \affiliation[3]{organization={Eindhoven University of Technology},
        city={Eindhoven},
        postcode={5612 AZ}, 
        country={The Netherlands}}
    
    \affiliation[4]{organization={Stanford University},
        city={Stanford},
        postcode={94305}, 
        state={CA},
        country={USA}}
        
    
    \begin{abstract} Question answering (QA) is the task of answering questions posed in natural language with free-form natural language answers extracted from a given passage. In the OpenQA variant, only a question text is given, and the system must retrieve relevant passages from an unstructured knowledge source and use them to provide answers, which is the case in the mainstream QA systems on the Web. QA systems currently are mostly limited to the English language due to the lack of large-scale labeled QA datasets in non-English languages. In this paper, we show that effective, low-cost OpenQA systems can be developed for low-resource{\REVISIONCOLOR\ contexts}. The key ingredients are (1) weak supervision using machine-translated labeled datasets and (2) a relevant unstructured knowledge source in the target language {\REVISIONCOLOR\ context}. Cut this word due to the length restriction of the submission system for abstracts
    Furthermore, we show that only a few hundred gold assessment examples are needed to reliably evaluate these systems. We apply our method to Turkish as a challenging case study, since English and Turkish are typologically very distinct{\REVISIONCOLOR\ and Turkish has limited resources for QA}. We present \SQUADTR, a machine translation of SQuAD2.0, and we build our OpenQA system by adapting \COLBERTQA\ and retraining it over Turkish resources and \SQUADTR{\REVISIONCOLOR\ using two versions of Wikipedia dumps spanning two years}. We obtain a performance improvement of{\REVISIONCOLOR\ 24--32\%} in the Exact Match~(EM) score and{\REVISIONCOLOR\ 22--29\%} in the F1 score compared to the \mbox{BM25-based} and \mbox{DPR-based} baseline QA reader models. Our results show that \SQUADTR makes OpenQA feasible for Turkish, which we hope encourages researchers to build OpenQA systems in other low-resource languages.  We make all the code, models, and the dataset publicly available at \url{https://github.com/boun-tabi/SQuAD-TR}. 

    \end{abstract}

    \begin{keywords}
    Question Answering \sep Open Domain Question Answering \sep OpenQA \sep Low-Resource Languages \sep Machine Translation
    \end{keywords}

    \maketitle

    \section{Introduction}
    \label{sec:intro}

    Question answering (QA) is the task of answering questions posed in natural
    language with free-form natural language answers. In its standard formulation,
    QA is posed in a highly constrained way. The system is given a passage and a question
    with a guarantee that the answer can be found in the passage~\citep{rajpurkar-etal-2016-squad,kwiatkowski-etal-2019-natural,joshi-etal-2017-triviaqa}.
    The main component of standard QA systems is a \emph{reader}, which takes a passage
    and a question as input and returns an answer. Present day systems are
    extremely successful at such tasks, often surpassing human performance~\citep{bui-etal-2020-state}.
    However, they are of limited use, since real-world question answering scenarios
    mostly do not involve gold passages or provide answerability guarantees.

    This observation has motivated a move towards Open Domain Question Answering (OpenQA),
    where only the question text is given as input without any passage. Related passages
    are retrieved from a large corpus by a \emph{retriever} and then used by the
    reader to predict an answer. In this setting, there is no guarantee that the
    retrieved passages will contain the answer, and the success of the system thus
    depends on having a successful retriever module to provide appropriate passages
    to the reader.

    Recent years have seen rapid improvements in such systems stemming from the use
    of neural retriever modules that can provide semantically rich representations
    of documents. We are approaching the point where OpenQA systems will be as
    effective as standard QA systems \citep{khattab-etal-2021-relevance}.

    However, this rapid progress in both standard QA and OpenQA systems is largely
    confined to English {\REVISIONCOLOR and high-resource language contexts}. Progress
    in other languages {\REVISIONCOLOR and low-resource scenarios} is constrained by
    a scarcity of gold data. While there are some high-quality multilingual
    resources in this domain \citep{lewis-etal-2020-mlqa, artetxe-etal-2020-cross},
    the amount and diversity of such data remain low. The cost of creating new datasets
    is the main obstacle to progress in this area {\REVISIONCOLOR for low-resource language contexts.

    We use the phrase \emph{``low-resource language context''} to refer to \textit{any language} as well as \textit{any domain}---such as e-commerce~\cite{shen-etal-2023-xpqa}, medical~\cite{pergola2021boosting,daniel-etal-2019-towards}, legal~\cite{ghosh-etal-2023-dale}, finance~\cite{sun:kdd23ws}, customer service~\cite{zheng-etal-2023-dialogqae}, space~\cite{darm-etal-2023-discosqa}---wherein the available OpenQA data is scarce. In other words, the low resource status of a context depends both on the language and the domain: a high resource language with ample resources, like English, may also face data limitations within certain domains or scenarios.

    Several effective methods have been proposed for overcoming various challenges and data limitations when building OpenQA systems for low-resource language contexts~\cite{shen2022lowresource}. However, most current research primarily focuses on low-resource language contexts in English~\cite{shen2022lowresource}, emphasizing the need for analyzing the efficiency and effectiveness of these methods when applied to low-resource contexts in non-English languages. }

    In this paper, we address the following research question: \emph{Can we build
    cost-effective QA systems for low-resource {\REVISIONCOLOR language contexts}
    without having a gold training dataset?} As a positive answer to this question,
    we propose a cost-effective approach to remedying the data scarcity problem
    for the QA task in non-English languages{\REVISIONCOLOR. This serves as a use case for low-resource language contexts}.
    Our proposal extends previous work on standard QA in languages other than
    English \citep{mozannar-etal-2019-neural,dhoffschmidt-etal-2020-fquad, abadani2021parsquad},
    and we argue that adopting the OpenQA formulation of the problem is a key step
    {\REVISIONCOLOR to remedy data-scarcity issue for QA applications in non-English languages}.
    For OpenQA, only gold question--answer pairs {\REVISIONCOLOR are required},
    and only for assessment. In particular, passages need not be a component of
    the gold data, since they are retrieved by the system to use as (perhaps noisy)
    evidence. Our formulation still requires training data, but this can be created
    by automatic translation from English datasets. These translations may contain
    mistakes, but we show that they can still lead to robust QA systems. Whereas
    the cost of creating a dataset like SQuAD~\citep{rajpurkar-etal-2016-squad, rajpurkar-etal-2018-know}
    can be upwards of US\$50,000, our costs are only around US\$500, most of which
    is for machine translation services. The cost of creating a gold assessment
    set could in principle be very large, but we show that one can get robust assessments
    of OpenQA systems with only around 200 question--answer pairs. Such gold datasets
    can be created by a small team very quickly.

    {\REVISIONCOLOR

    We make several key contributions to the field of OpenQA in low-resource language contexts:

    \begin{itemize}\item We demonstrate that QA is feasible in low-resource language contexts with the OpenQA formulation, even without any gold labels for training. This finding has significant implications for the accessibility and scalability of OpenQA systems in low-resource language contexts.

    \item We provide in-depth qualitative and quantitative analyses on the efficiency and effectiveness of OpenQA systems in non-English contexts, particularly when using noisy labels obtained through machine translation, offering insights into potential improvements of these models.

    \item We show that only a few hundred gold assessment examples are needed to effectively evaluate OpenQA systems, significantly reducing the resources and time required for model evaluation in various language settings.

    \item Our results highlight that increasing the size of unstructured knowledge sources can have varying effects on the performance of OpenQA systems, depending on the ability of the retriever systems to manage noise effectively.

    \item We release \SQUADTR, a large-scale Turkish Question Answering dataset, that was obtained by automatically translating SQuAD2.0. This resource facilitates building efficient and effective OpenQA systems in Turkish and serves as an example for other languages.\end{itemize}

    The rest of the paper is organized as follows: Section~\ref{sec:related_work_background} reviews related work for context and background. Section~\ref{sec:methods} outlines datasets, models, and methods with a focus on transparency and reproducibility. Section~\ref{sec:experiment_results} presents results, emphasizing key findings. Section~\ref{sec:discussion} discusses the results and their implications. Section~\ref{sec:generalization} shares key parameters to adapt and generalize our approach across diverse low-resource language contexts. Finally, Section~\ref{sec:conclusion} concludes with a summary and potential future directions.

    }
    \section{Related Work}
    \label{sec:related_work_background}
    \subsection{English Question Answering Datasets}

    There are many QA datasets for English used to address different challenges; see
    \citet{cambazoglu2021review} for a thorough review. One class of QA datasets consists
    of multiple-choice questions. MCTest~\citep{richardson-etal-2013-mctest} is an
    early dataset built in this style (see also CBT \citep{hill2015goldilocks};
    Booktest~\citep{bajgar2016embracing}). MCTest contains 2640 human-generated questions
    associated with a correct answer from a set of candidate answers. The
    questions and answers are based on 660 short fictional stories at a grade-school
    level. The fictional nature of the stories limits the use of world knowledge to
    answer the questions, which is one of the special challenges of this dataset. The
    main drawback of MCTest is its small size.

    SQuAD1.0~\citep{rajpurkar-etal-2016-squad} was the first major extractive reading
    comprehension dataset. SQuAD1.0 contains over 100K examples, and each example
    is a question--passage--answer triple, where annotators selected a span of text
    from the passage as the answer to the question. SQuAD2.0~\citep{rajpurkar-etal-2018-know}
    is a follow-up that includes over 50K additional examples representing
    unanswerable questions. The goal here is to encourage the development of systems
    that detect whether a question is answerable based on the passage given and abstain
    from answering if necessary~\citep{kamath-etal-2020-selective}. Although we did
    not use unanswerable questions in our experiments and they are out of the scope
    of this paper, we built \SQUADTR from SQuAD2.0 to facilitate future research on
    unanswerable questions in Turkish.

    HotPotQA~\citep{yang-etal-2018-hotpotqa} extends the extractive reading
    comprehension paradigm to multi-hop questions, i.e., questions whose answers need
    to be pieced together from information in multiple passages. A closely related
    task is multi-hop claim verification, as in HoVer~\citep{jiang-etal-2020-hover}.

    Another class of datasets leverages an existing set of human-generated
    question--answer pairs, and augments these with supporting passages from
    external knowledge sources. A prominent example of this type of dataset is TriviaQA~\citep{joshi-etal-2017-triviaqa},
    which contains 95K question--answer pairs that were prepared by trivia
    enthusiasts. The question--answer pairs are accompanied by documents retrieved
    from the Web and Wikipedia. In a similar manner, \citet{dunn2017searchqa} built
    SearchQA by using the Google search engine to retrieve context snippets relevant
    for question--answer pairs obtained from the Jeopardy!\ game show archive.\footnote{\url{http://j-archive.com}}

    Search engine query logs are also used as a source of examples. WikiQA~\citep{yang-etal-2015-wikiqa}
    and Natural Questions~\citep{kwiatkowski-etal-2019-natural} are the most commonly
    used datasets in this class. WikiQA is derived from the 3K most frequent user
    queries in the query logs of the Bing search engine. Each query is paired with
    a Wikipedia page clicked by at least five unique users. If a sentence in the
    summary part of the associated Wikipedia page includes the answer to the query,
    the sentence is labeled as \textit{correct}, otherwise \textit{incorrect}. This
    version of the QA task is referred as \textit{answer-sentence selection}, as it
    only selects the target sentence answering the question without requiring extraction
    of the correct answer span from that sentence. WikiQA includes question--page pairs
    with no \text{correct} sentences, so the dataset can also be used to build
    \textit{answer triggering} models, which predict whether the sentences include
    the answer or not and then select a sentence only if it answers the question.

    Like WikiQA, Natural Questions~(NQ) relies on queries to a real search engine.
    NQ contains a total of 320K examples with queries obtained from Google query
    logs. Each query is associated with a Wikipedia page, which may or may not
    contain the answer for the query. If the Wikipedia page has the answer,
    \textit{a long answer} is included in the example to show the passage answering
    the question. The example may also contain \textit{a short answer} denoting the
    short form of the target answer. If the example contains neither a long nor a
    short answer, then no answer span exists on the page. NQ is a challenging
    dataset with realistic queries supported by high-quality annotations for the
    long and short answers. Like WikiQA, NQ also provides an opportunity to build
    answer triggering models with its examples having no long and short answers.

    All of these datasets can be re-cast in the OpenQA mould, assuming we can find
    a large collection of relevant unlabeled documents to be used as a knowledge source.
    SQuAD1.0, NQ, TriviaQA, and HotPotQA have been extensively explored in these
    terms~\citep{khattab-etal-2021-relevance,lee-etal-2019-latent}.

    \subsection{Multilingual Question Answering}
    \label{sec:related:multi}

    Various methods have been used to address the dataset bottleneck for QA in non-English
    languages~\citep{chandra2021survey}. One approach is to curate in-language
    datasets from scratch. A number of datasets for different languages have been created
    in this way. We provide a summary in Table~\ref{tab:multilingual_qa_datasets}.
    Datasets created in this manner are likely to be of high quality, but they are
    expensive, {\REVISIONCOLOR labor-intensive,} and time-consuming to create.

    \begin{table*}
        [tp]
        \centering
        \setlength{\tabcolsep}{12pt}
        \begin{tabular}{l l c}
            \toprule \textbf{Dataset}                       & \textbf{Language}         & \textbf{Number of examples} \\
            \midrule KorQuAD~\citep{lim2019korquad1}        & Korean                    & 70,079                      \\
            FQuAD~\citep{dhoffschmidt-etal-2020-fquad}      & French                    & 62,003                      \\
            SberQuAD~\citep{efimov2020sberquad}             & Russian                   & 50,364                      \\
            CMRC 2018~\citep{cui-etal-2019-span}            & Chinese                   & 19,071                      \\
            GermanQuAD~\citep{moller-etal-2021-germanquad}  & German                    & 13,722                      \\
            \REVISIONCOLOR VIMQA~\citep{le-etal-2022-vimqa} & \REVISIONCOLOR Vietnamese & \REVISIONCOLOR 10,047       \\
            ARCD~\citep{mozannar-etal-2019-neural}          & Arabic                    & \phantom{0}1,395            \\
            \bottomrule
        \end{tabular}
        \caption{QA datasets for non-English languages.}
        \label{tab:multilingual_qa_datasets}
    \end{table*}

    One way to reduce the cost of creating in-language QA systems is to try to
    rely on the zero-shot transfer capabilities of cross-lingual models. In this approach,
    multilingual language models (e.g., mBERT~\citep{devlin-etal-2019-bert}; XLM~\citep{conneau2019cross};
    XLM-RoBERTa~\citep{conneau-etal-2020-unsupervised}) are finetuned on English QA
    datasets and then used to answer questions in target non-English languages.
    Multilingual models are efficient and cost effective, especially in large-scale
    applications requiring multiple language support. However, their performance on
    the target languages is relatively lower than models that use in-language embeddings~\citep{lewis-etal-2020-mlqa, dhoffschmidt-etal-2020-fquad, lim2019korquad1, moller-etal-2021-germanquad, budur-etal-2020-data}.

    Another cost-effective approach is to rely on machine translation services. In
    this approach, in-language training datasets are automatically obtained by
    translating an existing English dataset using machine translation (MT). Previously,
    SQuAD1.0~\citep{rajpurkar-etal-2016-squad} was translated into Arabic~\citep{mozannar-etal-2019-neural},
    French~\citep{dhoffschmidt-etal-2020-fquad}, and Spanish~\citep{carrino-etal-2020-automatic},
    and SQuAD2.0~\citep{rajpurkar-etal-2018-know} was translated into Persian~\citep{abadani2021parsquad}.
    Similar techniques have also been used in other areas of NLP~\citep{budur-etal-2020-data,xue-etal-2021-mt5}.
    {\REVISIONCOLOR For example, \citet{senel-etal-2024-kardes} recently introduced KardeşNLU\footnote{The term ``KardeşNLU'' translates to ``SiblingNLU'' in English.} using MT systems and Turkish resources in their process to obtain a cost-effective evaluation benchmark dataset for various Natural Language Understanding (NLU) tasks in other Turkic languages, which are often relatively \mbox{less-resourced} than Turkish in several NLP tasks. MT systems can also be effectively applied to extremely low-resource languages, including endangered Indigenous Languages~\cite{ebrahimi-etal-2023-findings, mager-etal-2023-neural, chen-abdul-mageed-2023-improving, shode-etal-2023-nollysenti, le-sadat-2021-towards-low}.}

    Using MT systems is undoubtedly productive, but relying on automatic
    translations for system assessment raises concerns about the validity of those
    assessments. To the extent that there are systematic errors in the MT output,
    assessment numbers are likely to be untrustworthy. To address this, \citet{lewis-etal-2020-mlqa}
    proposed MLQA, which is a \mbox{multi-way} aligned QA dataset to be used for
    evaluation purposes in 7 languages, with over 5K examples for each language.
    Similarly, \citet{artetxe-etal-2020-cross} developed \XQUAD, which consists of
    a subset of the SQuAD1.0 development dataset with human translations into 10
    languages, including Turkish. In what follows, we rely on the Turkish portion of
    \XQUAD, namely \XQUADTR, for evaluation as it is the only standard QA evaluation
    dataset that supports Turkish.

    \subsection{Open Domain Question Answering}

    Various methods are employed by researchers to develop OpenQA systems. An investigation
    of these methods can be found in the comprehensive survey presented by \citet{zhu2021retrieving}.
    {\REVISIONCOLOR Additionally, \citet{zhang-etal-2023-survey-efficient} provide a thorough analysis of different OpenQA systems, examining them in terms of complexity, efficiency, speed, resource demands, and other relevant factors.}
    Traditionally, OpenQA systems involve two pipelined components: a \textit{retriever}
    and a \textit{reader}. Given a question, the retriever is expected to retrieve
    candidate passages, and the reader is supposed to extract the target answer
    span from those retrieved passages.

    BM25 was a common choice for the retriever component in the earliest OpenQA
    systems~\citep{robertson1995okapi} and it remains in wide use today~\citep{nogueira-etal-2020-document, li2020parade, pradeep2021expando, WabiQA}.
    BM25 and other retrievers in its class rely on lexical matching. The guiding
    idea behind more recent neural retrievers is that lexical matching alone is
    not sufficiently semantic in nature to capture the nuanced ways in which
    passages can be relevant to user queries. Prominent recent examples of these neural
    retrievers include ORQA~\citep{lee-etal-2019-latent}, REALM~\citep{pmlr-v119-guu20a},
    DPR~\citep{karpukhin-etal-2020-dense}, RAG~\citep{lewis-et-al-2020-RAG}, ColBERT~\citep{khattab2020colbert},
    and SPLADE \citep{formal2021splade}. The leaderboards for OpenQA systems are currently
    dominated by systems that employ neural retrievers, though BM25 remains a very
    strong baseline, especially where latency and cost are important additional considerations
    beyond accuracy-style metrics \citep{santhanam-etal-2023-moving}.

    Neural retrievers, despite their advantages, suffer from the drawback of
    requiring a significant amount of storage space, especially for their indexes.
    To mitigate this limitation, \citet{yang-seo-2021-designing} proposed a
    solution that involves several techniques. These techniques include filtering out
    unnecessary passages prior to the retrieval step, consolidating retriever--reader
    models with a single encoder, and employing post-training compression (see also
    \citep{santhanam-etal-2022-colbertv2,Santhanam:Khattab-etal:2022}).

    The English-centric nature of research in this area is arguably holding back retriever
    development as well. The largest and most widely used dataset in this space is
    the \mbox{MS MARCO} Passage Ranking dataset~\citep{DBLP:conf/nips/NguyenRSGTMD16, craswell2020overview},
    and it contains only English texts and queries. However, \citet{bonifacio2021mmarco}
    translated MS~MARCO into 13 different languages using automatic translation.
    The result is mMARCO~\citep{bonifacio2021mmarco}, the first multilingual MS~MARCO
    variant. mMARCO has enabled much new research on multilingual passage retrieval.
    However, mMARCO does not have any labels within the text to denote answer spans,
    and so it cannot by itself support the development of multilingual QA systems.

    Neural information retrieval (IR) systems can begin from pretrained
    multilingual embeddings, and this can facilitate multilingual retrieval work.
    For example, \citet{asai-etal-2021-xor} use DPR in the retriever step and
    propose a cross-lingual transfer method (\mbox{XOR QA}) to obtain answers for
    unanswerable questions in the non-English languages from English Wikipedia. In
    order to do that, they (1) translate the questions in non-English languages
    into English, (2) find relevant passages and answer spans from English Wikipedia,
    and (3) translate the English answer spans back to the original language. The leaderboard
    for their paper, XOR-TyDi~\citep{asai-etal-2021-xor}, includes cross-lingual
    retrieval and OpenQA tasks.\footnote{\url{https://nlp.cs.washington.edu/xorqa/}}
    XOR-TyDi has similar motivations to our work, in that it tackles issues around
    building OpenQA systems in \mbox{non-English} languages effectively, but it
    differs from our work in substantive ways. To achieve its goals, XOR-TyDi
    makes the English knowledge source available to \mbox{non-English} languages
    with the help of a cross-lingual retriever. In contrast, we propose a method
    to build an in-language retriever that benefits from an existing in-language
    knowledge source in a \mbox{non-English} language. Both methods are based on translation,
    but our method benefits from translated data at training time (\texttt{TRANSLATE--TRAIN})
    even if the translation is noisy, whereas XOR-TyDi requires translations at
    test time (\texttt{TRANSLATE--TEST}), making the overall system highly
    vulnerable to translation errors.

    For the most part, the reader component in OpenQA systems is an extractive reader:
    given a retrieved passage and a question, it is trained to extract a substring
    of the passage corresponding to the answer. Readers of this sort are clearly
    best aligned with standard QA datasets where the answer is guaranteed to be a
    substring of the passage provided. In datasets where the answer can be expressed
    more indirectly, extractive strategies will fail. Extractive readers are also potentially
    sub-optimal for OpenQA systems, for two reasons: we might be able to retrieve multiple
    relevant passages, and the passages themselves might indirectly express the
    answer.

    The shortcomings of extractive readers were addressed in several works. \citet{lewis-et-al-2020-RAG}
    explore readers that can consume multiple passages and generate original texts
    in response. \citet{yu-etal-2022-kg} introduce KG-FiD, which incorporates knowledge
    graphs to rerank passages by utilizing Graph Neural Networks (GNN) before the reader
    generates the response.
    {\REVISIONCOLOR\citet{nie2019} present a multi-modal approach where the model is guided by heterogeneous knowledge sources and visual cues when generating responses within a conversational context. \citet{clinfo2024} adopt the Retrieval-Augmented Generation~(RAG)~\cite{lewis-et-al-2020-RAG} approach to generate answers using a large language model (LLM) for clinical questions based on medical literature in PubMed. \citet{mao-etal-2021-generation} compare the performance of both extractive and generative readers in an OpenQA system based on the passages obtained by a Generation-Augmented Retrieval~(GAR) step where additional context is generated for the queries to form generation-augmented queries. \citet{COT-2022} introduces \emph{Chain-of-Thought (CoT) Prompting}, which breaks the question into intermediate reasoning steps to generate a final answer. \citet{yao2023react} introduce \emph{ReAct} using CoT to create prompts that blend reasoning and suitable actions, such as seeking additional information from knowledge sources, for accurate and interpretable answers. \citet{khattab2022demonstrate} presents Demonstrate-Search-Predict~(DSP), a framework that orchestrates the retriever model and a language model generating a series of intermediary questions helping find multiple relevant passages that answer the question when combined. New toolkits like LangChain~\cite{Chase_LangChain_2022} and LlamaIndex~\cite{Liu_LlamaIndex_2022} have emerged to simplify the integration and orchestration of LLMs into multi-stage pipelines and external tools, where LLMs are guided by hand-crafted prompts for specific tasks including RAG, relying on in-context learning~\cite{McCann2018decaNLP, radford2019language, gpt3-2020}.  \citet{khattab2023dspy} introduce DSPy, a novel programming model and compiler that can eliminate reliance on hand-crafted prompts in LLM pipelines by automatically generating prompts and LLM invocation strategies based on a declarative program.}
    We leave exploration of generative readers for Turkish
    {\REVISIONCOLOR in the OpenQA formulation} for future work.

    Several end-to-end neural models have recently emerged in OpenQA (e.g., SOQAL~\citep{mozannar-etal-2019-neural};
    DPR \citep{karpukhin-etal-2020-dense}; ColBERT-QA~\citep{khattab-etal-2021-relevance};
    YONO~\citep{lee-etal-2022-need}). Early examples predominantly relied on
    sparse vector representations in the retrieval component. For instance, \citet{mozannar-etal-2019-neural}
    proposed SOQAL as an OpenQA system for Arabic using a hierarchical \mbox{TF-IDF}
    (Term Frequency-Inverse Document Frequency) retriever pipelined with a BERT-based
    reader~\citep{devlin-etal-2019-bert}. This was followed by an answer ranking component
    that assigns a score for each answer candidate obtained as a linear combination
    of the retriever and reader outputs. However, these retrievers, based on
    sparse representations, struggle to recognize similarity between synonyms and paraphrases
    that use different lexical terms.

    To address the sparse vector representation problem, \citet{karpukhin-etal-2020-dense}
    introduced DPR which is one of the early examples of dense retrievers in the OpenQA
    domain. DPR utilized dual-encoder architecture to encode dense and latent
    semantic representations of the questions and the contexts. Given a question,
    DPR was trained to distinguish the positive passages from the negative
    passages in the batch. One limitation of DPR was its use of a single vector for
    the question and the context, resulting in limited interactions between the
    terms in the two texts. \citet{khattab-etal-2021-relevance} recently developed
    \COLBERTQA as a novel end-to-end neural OpenQA model for English, offering more
    extensive and effective interaction between the question and context terms through
    a late-interaction mechanism. Alternatively, \citet{lee-etal-2022-need}
    proposed YONO, a single end-to-end architecture that jointly optimizes the retriever,
    reranking, and reader components. The fully end-to-end architecture of YONO
    contributes to its efficiency in terms of model size. However, there is a
    drawback to combining multiple components in a single architecture, as each component
    demonstrates different overfitting characteristics. This vulnerability becomes
    apparent especially when the training data is limited, which is often the case
    for low-resource languages.

    In this paper, we focus on two advanced end-to-end neural models used in
    OpenQA, the \mbox{DPR} and \COLBERTQA models, for their ability to provide
    dense representations of queries and passages. Each model is explained{\REVISIONCOLOR further in Section~\ref{sec:methods:models} along with the other models we use in the paper.}

    \section{Methodology}
    \label{sec:methods}

    {\REVISIONCOLOR In this section, we provide an overview of the datasets, models, and experimental settings used in this paper, aiming to enhance the transparency and reproducibility of our methodology and facilitate scrutiny of our findings.}

    \subsection{Datasets}
    \label{sec:datasets}

    {\REVISIONCOLOR\  In the following subsections, we outline the specifics of the data acquisition and preprocessing procedures we utilized to compile the datasets used in the experiments. }
    \subsubsection{SQuAD-TR}
    \label{sec:squad_tr}

    Inspired by previous work using machine translation as a stepping stone to obtain
    multilingual resources (\secref{sec:related:multi}), we translated SQuAD2.0~\citep{rajpurkar-etal-2018-know}
    to Turkish using Amazon Translate.\footnote{Amazon Translate was chosen thanks
    to the availability of AWS Cloud Credits for Research Grant for the authors, but
    it is possible to use other effective machine translation systems as well.
    Please refer to the disclaimer mentioned in the acknowledgements section for
    further information.} We translated the titles, context paragraphs, questions,
    and answer spans in the original dataset. As a natural consequence, we needed
    to remap the starting positions of the answer spans, since their positions were
    not maintained in the translated paragraphs. This is needed not only due to
    linguistic variation between the source and target languages~\citep{dhoffschmidt-etal-2020-fquad}
    but also because the translation task is inherently context dependent~\citep{mozannar-etal-2019-neural}.
    A text span may have totally different translations depending on its context.
    This is a challenging issue for obtaining consistent translations, particularly
    for Turkish due to the context-dependent morphological variation of Turkish words,
    as exemplified in Table~\ref{tab:automatic_post_processing_examples}. The problem
    with mapping all of the answer spans after translation is that it requires a substantial
    amount of time and manual work. However, it is still possible to recover part of
    them automatically, so we mapped the answer spans automatically in the target
    translations, as in much related work in different languages~\citep{mozannar-etal-2019-neural, dhoffschmidt-etal-2020-fquad, carrino-etal-2020-automatic, abadani2021parsquad}.

    In this automatic post-processing step, we first looked for spans of text in the
    context paragraph that exactly matched the answer text. If we found such a
    span, we kept that answer text along with its starting position in the translated
    text, following previous work~\citep{mozannar-etal-2019-neural, dhoffschmidt-etal-2020-fquad, carrino-etal-2020-automatic, abadani2021parsquad}.
    For answer texts without matching spans, we searched for the spans of text that
    approximately matched with the target answer text using character-level edit distance~\citep{levenshtein1966binary}.\footnote{We
    used the implementation in the Python \texttt{regex} package: \\
    \url{https://pypi.org/project/regex/2021.4.4}} We use different edit distance
    values based on the length of the answer text. For answer texts with lengths shorter
    than 4 characters, we try to match all spans that are 1-edit distance away
    from the answer text. For all other answer texts, we search for all spans that
    are \emph{up to} 3-edit distance away from the answer text and select all of the
    longest spans of texts that approximately match the target answer text. Table~\ref{tab:automatic_post_processing_examples}
    shows examples of the answer spans that are recovered as a result of this post-processing.

    This approximate matching is generally successful. However, for 25,528
    question--answer pairs in \SQUADTRTRAIN, neither exact nor approximate
    matching returns a span in the translated paragraph. We{\REVISIONCOLOR\ excluded}
    these question--answer pairs from \SQUADTRTRAIN~{\REVISIONCOLOR\ and made them available in a separate file}.
    This resulted in 259 paragraphs having no question--answer pairs. We{\REVISIONCOLOR\ excluded}
    those paragraphs from \SQUADTRTRAIN\ as well. Similarly, we{\REVISIONCOLOR\ excluded}
    3,582 question--answer pairs from the \SQUADTRDEV dataset, but we did not need
    to{\REVISIONCOLOR\ exclude} any paragraphs from \SQUADTRDEV, as all paragraphs
    had at least one question--answer pair where the answer text has a matching span
    in the paragraph.

    With this procedure, we obtained the training and evaluation splits of
    \SQUADTR, namely \SQUADTRTRAIN and \SQUADTRDEV, respectively. We used \SQUADTRTRAIN
    as a training dataset but did not use \SQUADTRDEV for evaluation in our research.
    We share it for future work. For evaluation, we instead used the Turkish split
    of \XQUAD~\citep{artetxe-etal-2020-cross}, namely \XQUADTR, which helped maximize
    the validity of our assessment results, since it is a high-quality, human-translated
    test set.

    Table~\ref{tab:dataset_stats} provides basic statistics of \SQUADTR and \XQUADTR
    along with the training and dev splits of the original SQuAD2.0 dataset (\SQUADEN),
    noted as \SQUADENTRAIN and \SQUADENDEV, respectively. The number of articles is
    identical for the \SQUADEN and \SQUADTR datasets, whereas the \SQUADTRTRAIN
    dataset has fewer paragraphs and answerable questions than \SQUADENTRAIN due
    to the excluded paragraphs and questions. Similarly, the \SQUADTRDEV dataset
    has fewer answerable questions than \SQUADENDEV, for the same reason. As a
    matter of course, the number of unanswerable questions did not change in any split
    of the \SQUADTR dataset, as the original unanswerable questions remain
    unanswerable after translation. We release \SQUADTR publicly.~\footnote{\SQUADTRURL}

    \begin{table*}
        [ht]
        \centering
        \setlength{\tabcolsep}{10pt}
        \resizebox{1\textwidth}{!}{%
        \begin{tabular}{@{} l l P{10cm} P{5cm} P{2.5cm} P{2.5cm} @{}}
            \toprule                                                                                                           & \multicolumn{1}{c}{\textbf{Language} } & \multicolumn{1}{c}{\textbf{Context span}}                                                                                                                                                                                                                                                                                                                                                                      & \multicolumn{1}{c}{\textbf{Question}}                                                                                           & \multicolumn{1}{c}{\textbf{\begin{tabular}[c]{@{}c@{}}Answer Text \\ (Before post processing)\end{tabular}}} & \multicolumn{1}{c}{\textbf{\begin{tabular}[c]{@{}c@{}}Answer Text \\ (After post processing)\end{tabular}}}   \\
            \midrule \multirow{2}{*}{\begin{tabular}[c]{@{}c@{}}\textbf{Example 1} \\ \textit{(Edit distance=1)}\end{tabular}} & Turkish                                & \multicolumn{1}{p{10cm}}{\ldots G\"{o}r\"{u}n\"{u}\c{s}\"{u}, o y\i{}lki MTV Video M\"{u}zik \"{O}d\"{u}lleri'nin MTV tarihinde en \c{c}ok izlenen yay\i{}n haline gelmesine ve \underline{\textbf{12.4 milyon}} izleyiciyi \c{c}ekmesine yard\i{}mc\i{} oldu;\ldots}                                                                                                                                          & \multicolumn{1}{p{5cm}}{2011 MTV M\"{u}zik \"{O}d\"{u}lleri'ni ka\c{c} ki\c{s}i izledi?}                                        & \multicolumn{1}{c}{\begin{tabular}[c]{@{}c@{}}12,4 milyon\\ \textit{(12.4 million)}\end{tabular}}            & \multicolumn{1}{c}{\begin{tabular}[c]{@{}c@{}}12.4 milyon\\ \textit{(12.4 million)}\end{tabular}}             \\
            \cmidrule{2-6}                                                                                                     & English                                & \multicolumn{1}{p{10cm}}{\ldots Her appearance helped that year's MTV Video Music Awards become the most-watched broadcast in MTV history, pulling in \underline{\textbf{12.4 million}} viewers;\ldots}                                                                                                                                                                                                        & \multicolumn{1}{p{5cm}}{How many people watched the 2011 MTV Music Awards?}                                                     & \multicolumn{1}{c}{12.4 million}                                                                             & \multicolumn{1}{c}{---}                                                                                       \\
            \hline
            \multirow{2}{*}{\begin{tabular}[c]{@{}c@{}}\textbf{Example 2} \\ \textit{(Edit distance=2)}\end{tabular}}          & Turkish                                & \multicolumn{1}{p{10cm}}{\ldots Kariyerindeki en uzun s\"{u}reli Hot 100 single'\i{} olma ba\c{s}ar\i{}s\i{}na ula\c{s}an \textquotedblleft{}Halo"un ABD'deki ba\c{s}ar\i{}s\i{}, Beyonc\'{e}'nin \underline{\textbf{2000'li}} y\i{}llarda di\u{g}er kad\i{}nlardan daha fazla listede ilk on single elde etmesine yard\i{}mc\i{} oldu.\ldots}                                                                 & \multicolumn{1}{p{5cm}}{Hangi on y\i{}l boyunca, Beyonce'\i{}n di\u{g}er kad\i{}nlardan daha fazla \c{s}ark\i{}s\i{} vard\i{}?} & \multicolumn{1}{c}{\begin{tabular}[c]{@{}c@{}}2000'ler\\ \textit{(2000s)}\end{tabular}}                      & \multicolumn{1}{c}{\begin{tabular}[c]{@{}c@{}}2000'li\\ \textit{(2000s)}\end{tabular}}                        \\
            \cmidrule{2-6}                                                                                                     & English                                & \multicolumn{1}{p{10cm}}{\ldots The album featured the number-one song "Single Ladies (Put a Ring on It)" and the top-five songs "If I Were a Boy" and "Halo". Achieving the accomplishment of becoming her longest-running Hot 100 single in her career, "Halo"'s success in the US helped Beyoncé attain more top-ten singles on the list than any other woman during the \underline{\textbf{2000s}}.\ldots} & \multicolumn{1}{p{5cm}}{For which decade, did Beyonce have more top ten songs than any other woman?}                            & \multicolumn{1}{c}{2000s}                                                                                    & \multicolumn{1}{c}{---}                                                                                       \\
            \hline
            \multirow{2}{*}{\begin{tabular}[c]{@{}c@{}}\textbf{Example 3} \\ \textit{(Edit distance=3)}\end{tabular}}          & Turkish                                & \multicolumn{1}{p{10cm}}{\ldots Amerika Kay\i{}t End\"{u}strisi Birli\u{g}i (RIAA), Beyonc\'{e}'yi 2000'lerin en iyi sertifikal\i{} sanat\c{c}\i{}s\i{} olarak toplamda \underline{\textbf{64 sertifikayla}} listeledi.\ldots}                                                                                                                                                                                 & \multicolumn{1}{p{5cm}}{2000'lerde ka\c{c} tane m\"{u}zik sertifikas\i{} ald\i{}?}                                              & \multicolumn{1}{c}{\begin{tabular}[c]{@{}c@{}}64 sertifikasyon\\ \textit{(64 certifications)}\end{tabular}}  & \multicolumn{1}{c}{\begin{tabular}[c]{@{}c@{}}64 sertifikayla\\ \textit{(with 64 certificates)}\end{tabular}} \\
            \cmidrule{2-6}                                                                                                     & English                                & \multicolumn{1}{p{10cm}}{\ldots The Recording Industry Association of America (RIAA) listed Beyoncé as the top certified artist of the 2000s, with a total of \underline{\textbf{64 certifications}}.\ldots}                                                                                                                                                                                                   & \multicolumn{1}{p{5cm}}{How many music certifications has she received in the 2000s?}                                           & \multicolumn{1}{c}{64 certifications}                                                                        & \multicolumn{1}{c}{---}                                                                                       \\
            \bottomrule
        \end{tabular}
        }
        \caption{Examples for the answer spans that are recovered in \SQUADTRTRAIN
        after the automatic post-processing steps.}
        \label{tab:automatic_post_processing_examples}
    \end{table*}
    \begin{table*}
        [ht]
        \centering
        \setlength{\tabcolsep}{10pt}
        \begin{tabular}{l l r r r r r}
            \toprule                                 &                  &                   &                     & \multicolumn{3}{c}{\textbf{Question Count}} \\
            \cmidrule{5-7} \textbf{Language}         & \textbf{Dataset} & \textbf{Articles} & \textbf{Paragraphs} & \textbf{Answerable}                        & \textbf{Unanswerable} & \textbf{Total} \\
            \midrule                                 & \SQUADENTRAIN    & 442               & 19035               & 86821                                      & 43498                 & 130319         \\
            \multirow{-2}{*}{English}                & \SQUADENDEV      & 35                & 1204                & 5928                                       & 5945                  & 11873          \\
            \midrule                                 & \SQUADTRTRAIN    & 442               & 18776               & 61293                                      & 43498                 & 104791         \\
                                                     & \SQUADTRDEV      & 35                & 1204                & 2346                                       & 5945                  & 8291           \\
            \cmidrule{2-7} \multirow{-3}{*}{Turkish} & \XQUADTR         & 48                & 240                 & 1190                                       & 0                     & 1190           \\
            \bottomrule
        \end{tabular}%
        \caption{Statistics for the \SQUADEN, \SQUADTR, and \XQUADTR datasets.}
        \label{tab:dataset_stats}
    \end{table*}

    \subsubsection{Knowledge Source}
    \label{sec:kb}

    As we discussed above, in OpenQA, evidence passages are not given to the reader
    along with the questions, but rather are retrieved from a large corpus. Thus,
    we first need to prepare a knowledge source containing the passages to be
    retrieved. We used the Turkish Wikipedia as the main part of our knowledge source.
    We obtained the passages in our knowledge base by extracting \emph{contents} and
    \emph{titles} from Turkish Wikipedia articles. However, we observed that the
    majority of the target information available in SQuAD2.0~\citep{rajpurkar-etal-2018-know}
    was not actually available in Turkish Wikipedia due to two main issues.

    One of these issues occurs when the article containing the target information
    in English Wikipedia is actually missing in Turkish Wikipedia. As an example,
    SQuAD2.0 has 50 question--answer pairs targeting 25 paragraphs about Canada's
    national public broadcaster \emph{CBC Television}\footnote{\url{https://en.wikipedia.org/wiki/CBC_Television}}
    referenced as an article in the English Wikipedia. However, there is no
    corresponding article for the same entity in Turkish Wikipedia but rather on
    \emph{TRT},\footnote{\url{https://tr.wikipedia.org/wiki/TRT}} \mbox{which is} \mbox{T\"{u}rkiye's
    national public broadcaster.} Therefore, all the information required to
    answer the questions about the Canadian CBC Television is missing in Turkish
    Wikipedia. Another issue happens when the target article is actually available
    in Turkish Wikipedia with information-rich content but is missing the target information
    due to cultural bias. For example, SQuAD2.0 dataset has a question \emph{When
    was the first known use of the word ``computer''?} targeting a passage in the English
    Wikipedia article \emph{Computer}.\footnote{\url{https://en.wikipedia.org/wiki/Computer}}
    The corresponding article \emph{Bilgisayar}\footnote{\url{https://tr.wikipedia.org/wiki/Bilgisayar}}
    in the Turkish Wikipedia\footnote{\REVISIONCOLOR\ \WIKITROLD.} does not have any
    information about the etymological origin of the English word `\emph{computer}',
    but instead the origin of its Turkish translation `\emph{bilgisayar}'. \citet{asai-etal-2021-xor}
    succinctly describe the issues behind these two examples as \emph{information
    scarcity} and \emph{information asymmetry}, which can be commonly called \emph{missing
    information} in the knowledge source of the target language.

    The missing information issues will probably resolve gradually as the Turkish
    Wikipedia grows over time in terms of the number of articles and their quality.
    {\REVISIONCOLOR\  However, it is worth noting that this expansion may also introduce noise into the system, particularly when new articles act as distractors for the questions.}
    To quantify the {\REVISIONCOLOR\ overall} effect of the {\REVISIONCOLOR\ expansion of}
    the knowledge source on the success of the OpenQA models, we used two
    different dumps of the Turkish Wikipedia with the dates spanning about 2 years,\footnote{We
    used the data dumps of May 31st, 2021 and May 1st, 2023} which we call
    \WIKITROLD and \WIKITRNEW.

    The missing information issues will understate the performance of the
    retriever models in the OpenQA systems if not mitigated properly. To mitigate these
    issues, we appended the target context passages of the \SQUADTRTRAIN and
    \XQUADTR~\citep{artetxe-etal-2020-cross} datasets to the Turkish Wikipedia articles~(\WIKITR)
    to complete our knowledge source. It should be noted that we do not append
    answer texts, but rather only the \emph{contexts} and \emph{titles}. In this
    way, we made the target passages in our knowledge source available to our models
    while ensuring the validity of our experimental protocol. As a result, the total
    number of passages in our knowledge source increased slightly with the addition
    of 19,117 unique passages in the \SQUADTRTRAIN and \XQUADTR datasets to the existing
    articles in the Turkish Wikipedia dump used.

    We split the combined passages of varying lengths in the knowledge source into
    equal chunks of passages using an enhanced whitespace tokenizer, as in the DPR
    model~\citep{karpukhin-etal-2020-dense}. The original DPR model for English
    segments the passages into 100-word chunks resulting in 142 tokens (subwords)
    on average when the BERT~\citep{devlin-etal-2019-bert} tokenizer is used.
    Turkish sentences produce about 1.3 times longer sequence of tokens with the
    same number of words, when the BERTurk~\cite{stefan_schweter_2020_3770924} tokenizer
    is used. The longer sequence of tokens in Turkish sentences can be attributed
    to the very rich suffixing morphology of Turkish. For this reason, unlike
    \citet{karpukhin-etal-2020-dense}, we split the passages into 75 words instead
    of 100 words, as 100-word segments in Turkish run a high risk of being truncated
    by the ColBERT model~\citep{khattab2020colbert}, which accepts up to 180 tokens
    for documents by default. After splitting the combined passages into equal
    chunks of 75 words, we obtained a total of 1.7M and 2.1M passages for the Wikipedia
    dumps dated 2021 and 2023, respectively.

    The resulting combined passages then served as the knowledge sources in our
    study. The basic statistics of these knowledge sources and the one used in the
    original DPR model for English are given in Table~\ref{tab:knowledge_source_stats}.

    \begin{table*}
        [ht]

        \centering
        \setlength{\tabcolsep}{3.2pt}
        \begin{tabular}{@{}lccclcc@{}}
            \toprule          &                   &                     &                         &                                & \multicolumn{2}{c}{\textbf{Passage Length}} \\
            \cmidrule(l){6-7} & \textbf{Language} & \textbf{Short Name} & \textbf{Wikipedia Date} & \textbf{Passage Count}         & \textbf{Word Count (Max)}                  & \textbf{Token Sequence Length (Avg)} \\
            \midrule DPR      & English           & \WIKIEN             & Dec 20, 2018            & \multicolumn{1}{r}{21,015,324} & 100                                        & 142                                  \\
            \midrule Ours     & Turkish           & \WIKITROLD          & May 31, 2021            & \multicolumn{1}{r}{1,719,277}  & 75                                         & 136                                  \\
                              &                   & \WIKITRNEW          & May 01, 2023            & \multicolumn{1}{r}{2,192,776}  & 75                                         & 136                                  \\
            \bottomrule
        \end{tabular}%

        \caption{Basic statistics for the knowledge sources used in our study and
        the one used in the \mbox{DPR model of \cite{karpukhin-etal-2020-dense}}.}
        \label{tab:knowledge_source_stats}
    \end{table*}

    { \REVISIONCOLOR\
\subsection{Models}\label{sec:methods:models} In this section, we share the background information about the retriever and reader models we used in our study. The same reader models are used for both the standard QA formulation and the OpenQA formulation in our experiments. }
    {\REVISIONCOLOR\
\subsubsection{Retriever Models}\label{sec:methods:openqa_models} \CUSTOMVSPACE \paragraph{Okapi BM25:}\label{sec:methods:bm25_model} The Okapi BM25 model, often abbreviated as BM25, is a probabilistic relevance algorithm that has been widely adopted for many years~\cite{robertson-bm25}. BM25 seeks to address core limitations of TF-IDF. For example, TF-IDF tends to be biased toward long documents. BM25 addresses this deficiency by incorporating document length normalization in addition to the conventional term frequency and inverse document frequency components.

    The BM25 algorithm is formally defined as follows. Let $q$ denote a query, $d$ a document, $q_{i}$ the $i$'th term of $q$, $f(q_{i}, d)$ the frequency of $q_{i}$ in document $d$, $|d|$ the length of document $d$, and $\textit{avgdl}$ the average document length in the document collection. BM25 has two hyperparameters: $k_{1} > 0$ adjusts the impact of term frequencies, and $0 \leq b \leq 1$ adjusts the document length penalty. The BM25 score for a document $d$ with respect to a given query $q$ is calculated as: \begin{equation}\label{eq:bm25}\resizebox{.9\hsize}{!}{$\textit{BM25}(q, d) = \sum\limits_{i=1}^{n}\text{IDF}(q_{i}) \cdot \dfrac{ f(q_i,d) }{ f(q_i,d) + k_1 \cdot \left( 1-b+b \cdot \dfrac{\lvert d \rvert} {\textit{avgdl}} \right) }$}\end{equation}
    \begin{align}\text{IDF}(q_{i}) = \log\frac{N}{\textit{df}_{q_i}}\end{align} where $N$ is the total number of documents in the collection and $\textit{df}_{q_i}$ is the number of documents in the collection containing the term $q_{i}$.

    In summary, the BM25 algorithm calculates a relevance score for a query $q$ and document $d$ based on the occurrence of the query terms in the document and the document collection, taking into account the document length to ensure a fair assessment.

    }

    \paragraph{DPR:}
    \label{sec:methods:dpr}

    The DPR model~\citep{karpukhin-etal-2020-dense} employs a BERT-based dual-encoder
    architecture for the retriever component within an end-to-end OpenQA system.{\REVISIONCOLOR\ DPR has two BERT-based encoders: one for queries (denoted as $E(q)$) and another for documents (denoted as $E(d)$). Unlike similar dual-encoder setups that share an embedding layer ~\citep{question-tagging2020}, DPR uses separate word embedding layers for each encoder. The DPR encoders extract the representation from the built-in \texttt{[CLS]} token and output a fixed-size vector representation (dimension $d = 768$ for the BERT-base models). These encoders are utilized within the retriever component of an end-to-end OpenQA system during the training and test time.

    The DPR encoders are trained by optimizing the negative log likelihood of the positive passage, $d^{+}$, against a batch of negative passages, $\mathbb{B}=\{d_{1}^{-}, d_{2}^{-}, \dots, d_{m}^{-}\}$ for a given query $q$ using the following loss function: \begin{align}\resizebox{.8\hsize}{!}{$\text{NLL}(q, d^{+}, \mathbb{B}) = -\log\frac{e^{\textit{\textit{score}}(q, d^+)}}{e^{\textit{score}(q, d^+)}+ \sum_{i=1}^{m} e^{\textit{DPR}(q, d^-_m)}}$}\end{align} where $\textit{DPR}(q, d)$ is defined as: \begin{equation}\label{eq:dpr_scoring_function}\textit{DPR}(q, d) = E(q) \cdot E(d)\end{equation} which measures similarity between the query and document vectors. After training, the passages are encoded using the document encoder and indexed offline. The scoring function in Eq.~\ref{eq:dpr_scoring_function} is also used for ranking the documents at inference time.

    }

    {\REVISIONCOLOR\ A lightweight retriever like BM25 provides a training bootstrap dataset for DPR training containing positive passages along with negative passages which are also referred as \textit{hard-negatives} by~\citet{karpukhin-etal-2020-dense}}.
    A crucial aspect of DPR involves learning the similarity between questions and
    passages by employing in-batch negatives{\REVISIONCOLOR\, distinct from hard-negatives}.
    These in-batch negatives are composed of relevant passages from other
    questions within the same training batch
    {\REVISIONCOLOR\ and employed alongside hard-negatives.} The primary strength
    of utilizing in-batch negatives lies in expanding the number of training
    examples effectively while keeping the memory footprint minimal. A clear strength
    of DPR is that it can encode all passages offline and store them in a fixed index,
    and query and document storing can be very fast. Its main weakness is that allows
    for only very minimal interactions between queries and documents.

    \paragraph{ColBERT-QA:}
    \label{sec:methods:colbert-qa}

    The ColBERT-QA system of \citet{khattab-etal-2021-relevance} is an OpenQA
    system built on top of the ColBERT retriever model~\citep{khattab2020colbert}.
    In ColBERT-QA, the retriever is iteratively fine-tuned using weak supervision from
    the QA dataset so that it can perform task-specific retrieval. ColBERT-QA
    standardly uses an extractive reader, though its fine-tuned retriever is compatible
    with a wide range of reader designs.

    The hallmark of the ColBERT model is its \textit{late interaction} mechanism: both
    queries and passages are separately encoded into sequences of token-level vectors
    corresponding roughly to the output states of a BERT encoder~\citep{devlin-etal-2019-bert}.
    Given a query $q$ encoded as a sequence of token-level vector representations $[
    q_{1}, \ldots, q_{m}]$ and a passage {\REVISIONCOLOR\  $d$} encoded as {\REVISIONCOLOR\ $[d_{1}, \ldots d_{n}]$},
    ColBERT computes
    {\REVISIONCOLOR\ the relevance score for $q$ and $d$ as \begin{equation}\label{eq:colbert_scoring}\textit{ColBERT}(q, d) = \sum_{i=1}^{m} \text{MaxSim}(q_{i}, d)\end{equation} where $\text{MaxSim}(q\textsubscript{i}, d)$ is defined as \begin{equation}\text{MaxSim}(q\textsubscript{i}, d) = \sum_{i}^{n} \max_{\{d_{j}\}_{j=1}^n}q_{i}\cdot d_{j}\end{equation} That is, we calculate}
    the similarity of every pair of vectors $q_{i}$ and{\REVISIONCOLOR\ $d_{j}$} and
    sum the scores only for the highest scoring{\REVISIONCOLOR $d_{k}$} for each
    $q_{i}$ (``MaxSim'').

    {\REVISIONCOLOR\
The scoring function serves a dual purpose, being used not only during the training process but also in testing. During training, ColBERT optimizes the cross-entropy loss in a binary classification task using the scores, $\textit{ColBERT}(q, d^{+})$ and $\textit{ColBERT}(q, d^{-})$, for the triple $(q, d^{+}, d^{-})$ where $q$ denotes the query, and $d^{+}$ and $d^{-}$ denote the positive and negative passages with respect to the query, respectively~\cite{khattab-etal-2021-relevance}.}
    For testing, this{\REVISIONCOLOR\ scoring function is} the basis for{\REVISIONCOLOR\ ranking}
    documents with respect to queries. The architecture allows all passages in the
    knowledge source to be encoded off-line and indexed for fast comparisons with
    query representations.

    As a pure retriever, ColBERT achieves state-of-the-art results across a wide variety
    of IR benchmarks~\citep{santhanam-etal-2022-colbertv2} and it can be
    implemented in a low-latency, space-efficient manner~\citep{Santhanam:Khattab-etal:2022}.
    ColBERT-QA is a powerful example of recent general-purpose approaches to OpenQA,
    and so we base our models on this architecture. To adapt the model to Turkish,
    we made only language-specific adjustments (\secref{sec:methods:openqa}).

    {\REVISIONCOLOR\
\subsubsection{Reader Models}\label{sec:methods:stadardqa_models} \CUSTOMVSPACE \paragraph{BERT:}\label{sec:methods:bert_model} BERT~\cite{devlin-etal-2019-bert} has emerged as a revolutionary NLP model, fundamentally altering how contextual understanding is achieved within sentences by employing Transformers~\cite{transformers} as its core architecture. One key landmark of BERT is that it is pretrained on large text corpora through self-supervision and adapts its representations to specific NLP tasks to deliver state-of-the-art results across a wide range of applications.

    During BERT's pre-training, it uses two main objectives. The Next Sentence Prediction (NSP) objective involves predicting if one sentence follows another. The Masked Language Modeling (MLM) objective involves randomly masking or replacing words in input sentences and training BERT to predict the original words. BERT generates a contextual representation for every input token. For answer-span extraction, we follow \citet{devlin-etal-2019-bert} in adding a span-classification head that predicts the start and end positions of the answer for a given question within a given passage.

    \paragraph{mBERT and BERTurk:}\label{sec:methods:mbert_berturk}

    Since its inception, various BERT variants have emerged to suit specific linguistic contexts. The multilingual BERT model (mBERT~\citep{artetxe-etal-2020-cross}) is trained on multilingual data, enabling it to handle multiple languages without fine-tuning on language-specific data. Additionally, there are language-specific BERT models, like BERTurk~\cite{stefan_schweter_2020_3770924} for Turkish, trained on extensive language-specific data to better capture nuanced characteristics of those languages. These variants showcase BERT's adaptability and versatility across diverse linguistic contexts, enhancing its utility for a wide range of natural language processing tasks.

    In addition to the data-driven adaptations, BERT has also undergone architectural modifications, resulting in improved variants such as RoBERTa~\citep{liu2019roberta}, XLM-RoBERTa~\citep{conneau-etal-2020-unsupervised}, ELECTRA~\citep{clark2019electra}, and DeBERTa~\citep{he2020deberta}.

    \paragraph{XLM-RoBERTa:}\label{sec:methods:xlm_roberta_model}

    XLM-RoBERTa~\cite{conneau-etal-2020-unsupervised} represents an extension of the RoBERTa model~\cite{liu2019roberta}, which is an optimized version of BERT designed to improve pre-training objectives and hyperparameters. The key distinction of RoBERTa~\cite{liu2019roberta} compared to BERT lies in its optimized pre-training approach. Unlike BERT, which employs static masking patterns and the NSP task during pre-training, RoBERTa utilizes dynamic masking and removes the NSP task. These modifications, among others, enable RoBERTa to generalize more effectively and outperform BERT on several downstream NLP tasks.

    XLM-RoBERTa takes this a step further by focusing on cross-lingual understanding. As a cross-lingual version of RoBERTa, it is trained to understand text in multiple languages. This is particularly valuable for multilingual applications where a single model is needed to handle text in different languages without fine-tuning on language-specific data. Its versatility and effectiveness make it a valuable tool for multilingual NLP applications, where understanding and processing text across different languages is essential. }

    {\REVISIONCOLOR\
\subsection{Standard QA and OpenQA Formulations}\label{sec:methods:experimental_protocol}

    In this section, we provide a detailed description of the methodologies and settings followed during the execution of our standard QA and OpenQA experiments as well as any other pertinent details necessary for the reproducibility and transparency of our research findings. }
    {\REVISIONCOLOR\
\subsubsection{Standard QA Formulation}\label{sec:methods:standardqa} }
    To help establish an upper-bound for OpenQA in Turkish, we first conducted a series
    of standard QA experiments. \citet{artetxe-etal-2020-cross} established a baseline
    for these experiments with an mBERT model~\citep{devlin-etal-2019-bert}
    trained on \SQUADENTRAIN~\citep{rajpurkar-etal-2018-know} and tested on \XQUADTR~\citep{artetxe-etal-2020-cross}
    as a crosslingual QA application. We extended this experiment in two ways.
    First, we changed the training dataset to \SQUADTRTRAIN while keeping all other
    aspects of \citeauthor{artetxe-etal-2020-cross}'s system fixed; the goal of this
    experiment is to begin to understand \SQUADTRTRAIN as a training resource.
    Second, we changed mBERT to BERTurk~\citep{stefan_schweter_2020_3770924} to
    see the effects of pairing an in-language model with an in-language dataset.{\REVISIONCOLOR\ Third, we substituted BERTurk with \XLMROBERTA~\cite{conneau-etal-2020-unsupervised} to understand the effect of this architectural improvement. }

    For these experiments, we finetuned the BERTurk, mBERT{\REVISIONCOLOR\, and \XLMROBERTA}
    models with the same hyperparameters using \SQUADTRTRAIN and \XQUADTR as the
    training and test datasets. We used a batch size of 16, without gradient
    accumulation, on a single NVIDIA Tesla V100 GPU. We applied $3 \times 10^{-5}$
    as the learning rate, used a maximum length of 384, with a document stride of
    size 128, and trained each model for 5 epochs using Huggingface's \texttt{transformers}
    library~\citep{wolf-etal-2020-transformers}, Version 4.14.0.dev0.\footnote{The
    choice of specific values for the hyperparameters in our study is primarily
    aimed at establishing an initial reference point for future studies within a
    constrained budget.}

    {\REVISIONCOLOR\ In all our reader models, we use standard evaluation metrics from the literature~\citep{rajpurkar-etal-2016-squad}}:
    Exact Match (EM) and F1 scores. EM is the percentage of the predicted answer
    texts matching at least one of the ground-truth answer texts in an exact
    manner. F1 is the average of the maximum overlap ratio between predicted answer
    tokens and ground truth answer tokens. While EM gives no credit to predictions
    that have no exact match in any of the ground truth answer texts, F1 gives partial
    credit to those predictions that have at least one partially matching ground truth
    answer token. We calculated the evaluation metrics on \XQUADTR as our test set.

    {\REVISIONCOLOR\
 \subsubsection{OpenQA Formulation} \label{sec:methods:openqa} }

    In this section, we turn to OpenQA for Turkish. We establish baselines using BM25
    and DPR~\citep{karpukhin-etal-2020-dense} as examples of sparse and dense
    retrievers. Then, we share the results of our proposed system based on ColBERT-QA~\citep{khattab-etal-2021-relevance}.
    We first review the main components of our system, the retriever and the reader.
    We conduct the experiments for each Wikipedia dump as the knowledge source
    separately, which allows us to observe the{\REVISIONCOLOR\ overall effect} of
    the growth in the knowledge source.

    \paragraph{Retriever}
    \label{sec:methods:openqa_retriever}

    \begin{figure*}
        \centering
        \includegraphics[width=\textwidth]{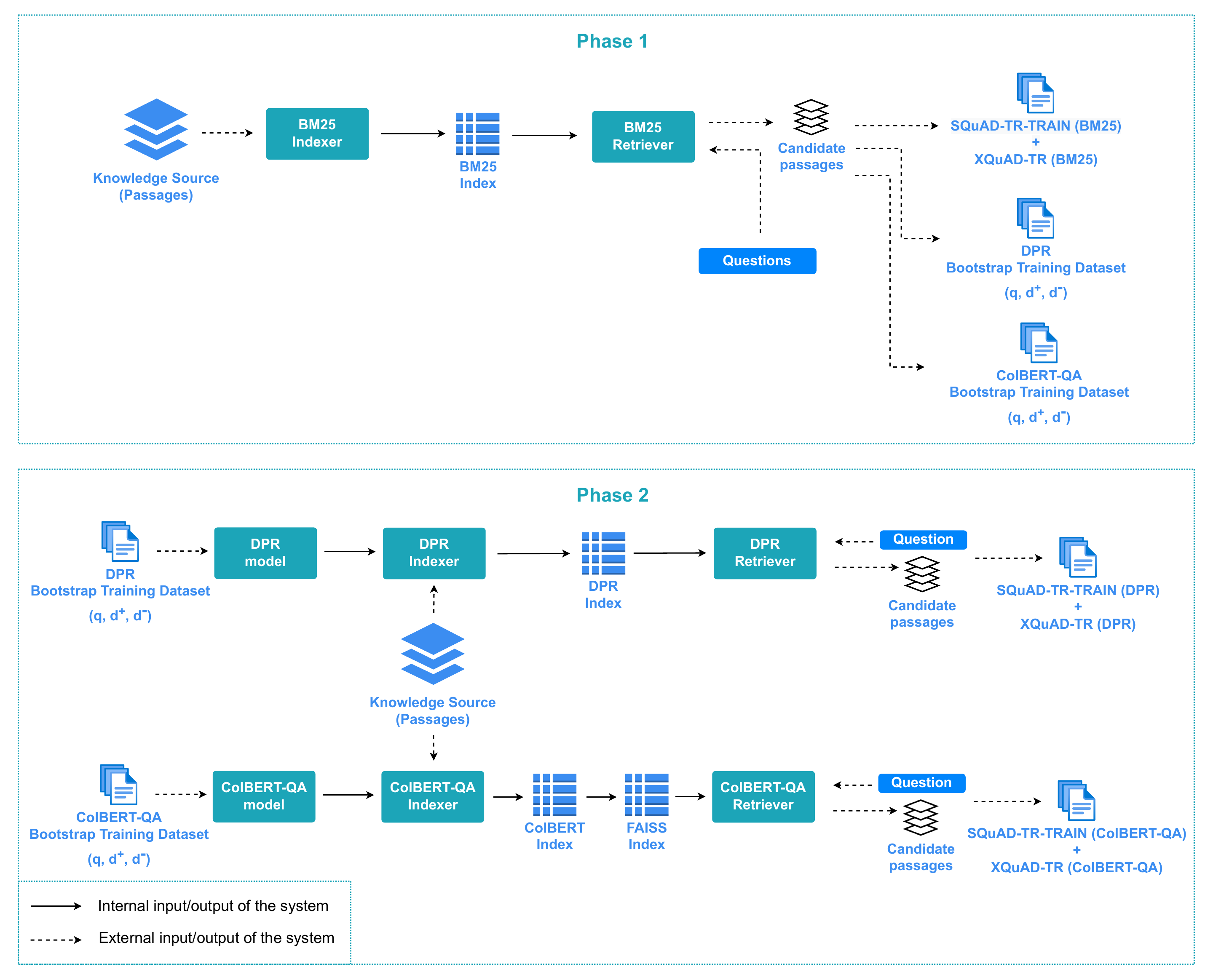}
        \caption{System overview diagram for the OpenQA retriever component. We assume
        the knowledge source is independent of the system for the sake of clarity in
        the figure.}
        \label{fig:open_qa_retriever_system_diagram}
    \end{figure*}

    The first step in building {\REVISIONCOLOR\ the DPR~\citep{karpukhin-etal-2020-dense} and}
    \mbox{ColBERT-based}~\citep{khattab2020colbert} retrievers
    {\REVISIONCOLOR\ in our experiments} involves a handful of steps that are
    specific to Turkish but that may have more general utility for cross-linguistic
    applications:
    \begin{enumerate}
        \item {\REVISIONCOLOR\ Both DPR and }ColBERT use the WordPiece tokenizer of
            the original BERT-base model~\citep{devlin-etal-2019-bert} in English. We
            replace these tokenizers with the WordPiece tokenizer of the BERTurk cased
            model~\citep{stefan_schweter_2020_3770924}, which was pretrained on a
            large Turkish corpus.

        \item The original tokenizer of the ColBERT model repurposes ``unused'' tokens
            in the tokenizer as query and document markers, which are available in the
            tokenizer of the original BERT-base model in English. As the BERTurk tokenizer
            did not have such unused tokens, we use alternative tokens for the
            document and question marker tokens. For the query marker we use a ``blush''
            emoji (U+1F60A) and for the document marker we use a ``smiley'' emoji (U+1F603),
            as they are unlikely to occur in the Wikipedia articles yet likely to be
            present in various non-English BERT models.

        \item {\REVISIONCOLOR\ Both DPR and} ColBERT{\REVISIONCOLOR\ originally}
            initialize their weights using those of the original BERT model in English.
            We use the BERTurk weights to initialize the{\REVISIONCOLOR\ DPR and}
            ColBERT weights before starting the training step. For languages without high-quality
            language-specific embeddings like BERTurk, one might use multilingual
            embeddings here instead.
    \end{enumerate}
    In light of the above steps, {\REVISIONCOLOR\ the resulting retrievers} might
    more properly be called {\REVISIONCOLOR\ the DPRTurk and} ColBERTurk
    retrievers. In the interest of clarity, we will continue to refer to them as
    {\REVISIONCOLOR\ the DPR and } ColBERT models.

    To train the retriever, we proceed in two phases, as outlined in Figure~\ref{fig:open_qa_retriever_system_diagram}.
    In phase~1 (top row of the figure), we build our baseline retriever models. We
    rely on BM25 to index our knowledge sources, using \texttt{pyserini}~\citep{Lin_etal_SIGIR2021_Pyserini}\footnote{\url{https://github.com/castorini/pyserini}}
    and \texttt{anserini}~\citep{lin2016toward}\footnote{\url{https://github.com/castorini/anserini}}
    wrappers for the Apache Solr search engine. We customize Apache Solr for Turkish
    by incorporating the Zemberek\footnote{\url{https://github.com/iorixxx/lucene-solr-analysis-turkish}}
    plugin~\citep{zemberek-solr-plugin} as a morphological stemmer for Turkish. In
    addition to the BM25 retriever, we use the DPR model to index our knowledge sources
    and build our baseline dense retriever.

    In our experiments, the BM25 retriever provides the bootstrap dataset to train
    {\REVISIONCOLOR\ DPR and} ColBERT-QA. With this lightweight BM25 retriever, we
    create a dataset of triples ($q, d^{+}, d^{-}$), where $q$ is a question in \SQUADTRTRAIN,
    $d^{+}$ is a positive passage containing the target answer span for $q$, and $d
    ^{-}$ is a negative passage that does not contain the target answer span for $q$.
    Both $d^{+}$ and $d^{-}$ are from the top $k$ results retrieved from the BM25
    index. More specifically, we create the dataset of triples by pairing every
    $d^{+}$ with every other $d^{-}$ where $d^{+}$ is from the top $k^{+}$ results
    and $d^{-}$ is from the top $k^{-}$ results ($k^{+} \leq k^{-}$) obtained for
    each question $q$.

    {\REVISIONCOLOR\ Specifically, we selected a maximum of three positive passages from the top $k=20$ results for both the DPR and \COLBERTQA models. In the case of DPR, we selected the top most negative passage out of top 20 results and used it as the hard-negative passage in a training example following~\citet{karpukhin-etal-2020-dense}. As for \COLBERTQA, we paired each positive passage with a negative passage from the top 100 results, per the method outlined by~\citet{khattab-etal-2021-relevance}}.

    For both of the Turkish Wikipedia dumps used as the knowledge source, the resulting
    bootstrap training dataset for the{\REVISIONCOLOR\ DPR model and } the
    \COLBERTQA model contains{\REVISIONCOLOR\ 64K} and 6M triples{\REVISIONCOLOR\, respectively,}
    for 86K questions, where $k^{+}=3$ and{\REVISIONCOLOR\ $k^{-}=20$}.{\REVISIONCOLOR\ It is worth mentioning that the DPR model amplifies the size of its training bootstrap dataset by incorporating the in-batch negatives during training. Additionally,}
    it can be noted that we could use all question--answer pairs in \SQUADTR that were
    originally labeled as answerable before translating SQuAD2.0. The reason for also
    including those question--answer pairs that we excluded from \SQUADTRTRAIN (\secref{sec:squad_tr})
    is that the retriever model, unlike the reader, does not require the location of
    the answer span in the context. Therefore, the retriever model can{\REVISIONCOLOR\ utilize}
    all question--answer pairs in \SQUADTRTRAIN.

    In phase~2 (bottom row of Figure~\ref{fig:open_qa_retriever_system_diagram}),
    we use our BM25-derived datasets to train a{\REVISIONCOLOR\ DPR model and} a \COLBERTQA
    model, and then we index our knowledge source using this retriever. These indexers
    compute the passage representations using{\REVISIONCOLOR\ DPR and }\COLBERTQA to
    project them into an embedding space where the question and passage representations
    are close to each other if the passage has an answer for the question.

    \begin{figure*}
        \centering
        \includegraphics[scale=.45]{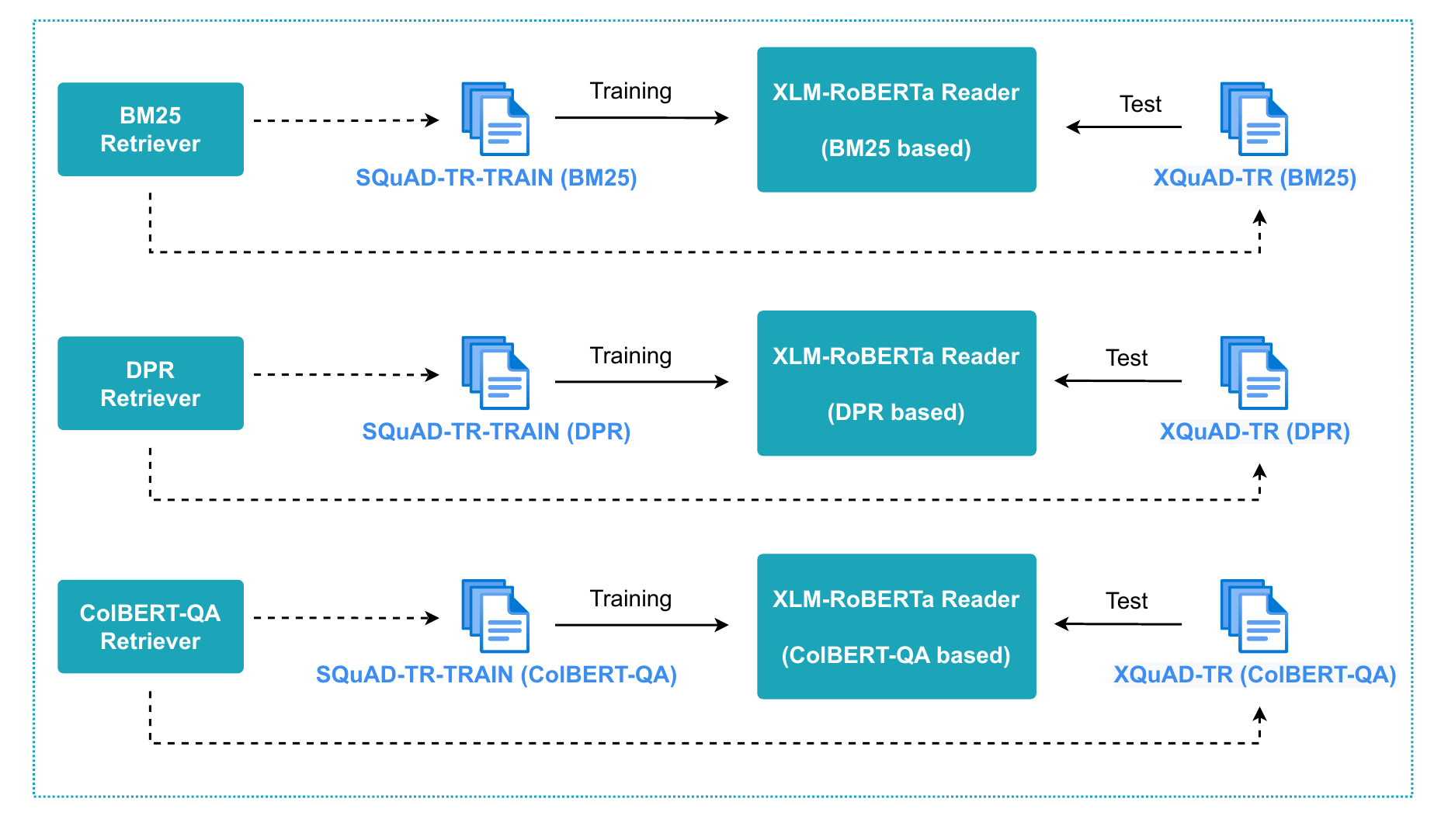}
        \caption{Overview diagram for the OpenQA reader component. For the sake of clarity
        in the figure, we assume the knowledge source that retrievers are based on
        is the same across all systems.}
        \label{fig:open_qa_reader_system_diagram}
    \end{figure*}

    {\REVISIONCOLOR\ We trained the DPR model for Turkish on 6 NVIDIA RTX A6000 GPUs in parallel. We used a batch size of 128 without gradient accumulation, aligning with the configuration yielding the best reported scores~\citep{karpukhin-etal-2020-dense}. Other parameters were set to their default values as specified in the original implementation. We indexed all the passages in the knowledge source, encoded by the resulting DPR model, using FAISS~\citep{johnson2019billion} in flat mode as the default configuration. }

    {\REVISIONCOLOR\ For training the} \COLBERTQA model in Turkish{\REVISIONCOLOR\, we used}
    a single {\REVISIONCOLOR\ NVIDIA RTX A6000 GPU} with a maximum document length
    of 180 and a batch size of 32 without gradient accumulation. Then, we indexed
    all the passages in the knowledge source once again, this time using \COLBERTQA.
    Following \citet{khattab-etal-2021-relevance}, we further reindexed \COLBERTQA
    indexed document embeddings using FAISS~\citep{johnson2019billion} in IndexIVFPQ
    mode with a 16384 partition and a sample rate of 0.3 to speed up the retriever
    component.

    Since there is no state-of-the-art model for OpenQA in Turkish yet, we compare
    the retriever and reader performance of our models with the performance of
    models based on the baseline BM25 and DPR retrievers. It is important to note that
    each retriever determines its own versions of the \SQUADTRTRAIN and \XQUADTR~\citep{artetxe-etal-2020-cross}
    datasets specific to that retriever and to the knowledge source it uses. For each
    retriever, we retrieve the top $k$ passages for each question in \SQUADTRTRAIN
    and \XQUADTR to set a context passage for that question
    {\REVISIONCOLOR\ from the top $k$ results}. {\REVISIONCOLOR\ In both the training and testing phases, we use the first positive result out of top 5 retrieved results. Should there be no positive result among these top 5 results, we resort to the first result, regardless of whether it is positive or negative.}

    The success of the retriever component sets an upper bound for the reader. Following
    previous works (DrQA \citep{chen-etal-2017-reading}; DPR~\citep{karpukhin-etal-2020-dense};
    REALM~\citep{pmlr-v119-guu20a}; \COLBERTQA~\citep{khattab-etal-2021-relevance}),
    we evaluated the success of the retriever component by means of \emph{Success@k},
    also noted as \emph{S@k}, which is the ratio of the questions having a positive
    passage among the top $k$ results retrieved from the index. We evaluated \emph{S@k}
    values for $k \in \{1,5,20\}$. We supplemented \emph{S@k} with another metric,
    \emph{Count@k}, also noted as \emph{C@k}, which is the average number of
    positive passages among the top $k$ results retrieved for each question.

    Turkish has different morphological characteristics from English, as it is an agglutinative
    language and has more morphological variants of each word. For this reason, the
    evaluation scores depend heavily on the tokenization scheme that is used when evaluating
    the results. Therefore, we used three different tokenization schemes: \textit{whitespace},
    \textit{morphological}, and \textit{enhanced whitespace}. Whitespace
    tokenization calculates the \emph{S@k} and \emph{C@k} values after splitting
    the retrieved passage and answer text into tokens whenever it finds a
    whitespace character. The morphological tokenization scheme segments the passage
    and answer texts into a list of stems by stripping all suffixes in all words before
    calculating the \emph{S@k} and \emph{C@k} values. The enhanced whitespace
    tokenization, which is the standard tokenizer of the DPR model, breaks the
    text into tokens not only when it encounters a whitespace character but also
    whenever it finds a list of predefined punctuations. We used the uncased version
    for all tokenization schemes to bring them in line with the output of our morphological
    tokenizer (Zemberek~\citep{Zemberek}), which was uncased out of the box.

    {\REVISIONCOLOR\
While the performance of the retriever models on the evaluation dataset (\XQUADTR) offers useful insights,
    a better assessment in the context OpenQA
    comes from evaluating the retrievers on the training dataset (\SQUADENTRAIN). This is because we leverage the retrievers' outputs from the training dataset to prepare the retriever-specific training datasets for the respective readers. It is important to emphasize that the performance of the retrievers on \SQUADTRTRAIN and \XQUADTR may vary based on their ability to handle the machine-translated texts and human-translated texts. Hence, we also assess the retrievers' performance on \SQUADTRTRAIN. }

    \paragraph{Reader}
    \label{sec:methods:openqa_reader}

    \begin{table*}
        [ht]
        \centering
        \setlength{\tabcolsep}{12pt}
        \begin{tabular}{l c c c}
            \toprule \textbf{Reader Model}                                   & \textbf{Training Dataset}    & \textbf{EM}          & \textbf{F1}          \\
            \midrule mBERT~\citep{artetxe-etal-2020-cross} - \emph{Baseline} & \SQUADENTRAIN                & --                   & 55.40                \\
            mBERT                                                            & \SQUADTRTRAIN                & 50.00                & 64.76                \\
            BERTurk                                                          & \SQUADTRTRAIN                & 51.17                & 67.78                \\
            \REVISIONCOLOR XLM-RoBERTa                                       & \REVISIONCOLOR \SQUADTRTRAIN & \REVISIONCOLOR 52.18 & \REVISIONCOLOR 68.63 \\
            \bottomrule
        \end{tabular}
        \caption{Reader results for the standard formulation of QA task evaluated on
        \XQUADTR.}
        \label{tab:standard_formulation_reader_results}
    \end{table*}

    In line with the three sets of OpenQA retrievers obtained in the retriever step
    for each knowledge source with different Wikipedia dumps, we build three sets
    of reader components in our OpenQA system in Turkish, depending on which retriever
    their training and test sets are based on. Figure~\ref{fig:open_qa_reader_system_diagram}
    summarizes this process. We used the {\REVISIONCOLOR\\XLMROBERTA~\cite{conneau-etal-2020-unsupervised} model}
    along with its original tokenizer for each reader, and we finetuned them using
    the retriever and knowledge source specific versions of \SQUADTRTRAIN, namely \SQUADTRTRAINBMYYYY
    {\REVISIONCOLOR\, \SQUADTRTRAINDPRYYYY} and \SQUADTRTRAINCOLBERTYYYY, where
    $\texttt{YYYY}\in \{2021, 2023\}$ denotes the year of the Wikipedia dump.

    We used the same hyperparameters as described in~\secref{sec:methods:standardqa}
    to train and evaluate the BERTurk model on the retriever-specific datasets as shown
    in Figure~\ref{fig:open_qa_reader_system_diagram}. We also calculated EM and
    F1 scores on the open versions of \XQUADTR~\citep{artetxe-etal-2020-cross} for
    each reader model.

    \paragraph{Subsampling Test Sets for OpenQA}
    \label{sec:methods:evaluation_with_subsampled_datasets}

    One of our central goals is to efficiently create OpenQA systems. Machine
    translation costs are already manageable and controlled. However, creating
    gold test sets can lead to unexpectedly high costs, especially if the goal is to
    have thousands or tens of thousands of examples. Thus, a key question for us is:
    \emph{How small can our test sets be?}

    To begin to address this question, we ran a series of experiments in which we subsampled
    the OpenQA test sets that we obtained using \WIKITRNEW.
    {\REVISIONCOLOR\ To create test subsets, we randomly picked samples in sizes of 100, 200, 500, and 1000 from the complete test datasets, alongside using the full test datasets. We then evaluated each retriever's performance on the subsampled questions of different sizes. Similarly, for the readers, we obtained prediction results of the readers on these subsampled examples to measure the readers' performance across subsets of various sizes. This process was repeated 20 times for each size of the subsampled reader datasets, aiming to maintain a diverse representation of the entire test set for each subset size, thereby ensuring the reliability of the findings derived from the evaluation process.}

    \section{Experimental Results}
    \label{sec:experiment_results} {\REVISIONCOLOR\
In this section, we report the experimental results from both standard QA and OpenQA formulations. Additionally, we offer an overview of the resource usage patterns of the models involved. }
    \subsection{Standard QA Results}
    \label{sec:exp:standard}

    Table~\ref{tab:standard_formulation_reader_results} summarizes the results of
    the experiments for standard QA, where we experiment with mBERT, BERTurk{\REVISIONCOLOR\, and \XLMROBERTA}
    as readers and \SQUADENTRAIN and \SQUADTRTRAIN as training data.{\REVISIONCOLOR\ The results show that \XLMROBERTA yields the highest scores, even outperforming an in-language model, BERTurk, when trained on an in-language dataset \SQUADTRTRAIN.}
    The largest performance gap occurs when \SQUADENTRAIN is replaced with \SQUADTRTRAIN,
    indicating that in-language datasets are essential for high-performing
    standard QA models, even if the datasets are machine-translated and potentially
    noisy. The use of an in-language model instead of a multi-lingual one has a
    smaller positive impact on the performance
    {\REVISIONCOLOR\ when the model architecture remains the same}. This aligns with
    other recent findings in the literature~\citep{budur-etal-2020-data,liu-etal-2019-xqa}.

    \subsection{Open QA Results}
    \label{sec:exp:openqa_results} \CUSTOMVSPACE
    \subsubsection{Retriever Results}

    \begin{table*}
        [ht]
        \centering
        \setlength{\tabcolsep}{3.5pt}
        \resizebox{1\textwidth}{!}{%
        \begin{tabular}{l c ccc ccc ccc}
            \toprule                                                                        &                           & \multicolumn{3}{c}{\textbf{Whitespace}}     & \multicolumn{3}{c}{\textbf{Morphological}}  & \multicolumn{3}{c}{\textbf{Enhanced Whitespace}} \\
            \cmidrule(l){3-5} \cmidrule(l){6-8} \cmidrule(l){9-11} \textbf{Retriever Model} & \textbf{Knowledge Source} & \textbf{S@1/\RCONE}                         & \textbf{S@5/\RCFIVE}                        & \textbf{S@20/\RCTWENTY}                          %
                                                                                            & \textbf{S@1/\RCONE}       & \textbf{S@5/\RCFIVE}                        & \textbf{S@20/\RCTWENTY}                      %
                                                                                            & \textbf{S@1/\RCONE}       & \textbf{S@5/\RCFIVE}                        & \textbf{S@20/\RCTWENTY}                      \\
            \midrule BM25 - \emph{Baseline - Sparse}                                        & \WIKITROLD                & 42.79/0.43                                  & 58.91/\underline{0.85}                      & 66.64/1.17                                      & 45.97/0.46                                  & 62.27/0.95                         & 69.92/1.39                                  & 56.30/0.56                                  & 73.53/1.11                         & 82.10/1.55                                  \\
                                                                                            & \WIKITRNEW                & 41.68/0.42                                  & 58.15/0.85                                  & 66.22/1.18                                      & 44.62/0.45                                  & 61.43/\underline{0.95}             & 69.66/1.42                                  & 55.21/0.55                                  & 72.52/1.10                         & 81.18/\underline{1.55}                      \\
            \midrule

DPR - \emph{Baseline - Dense}                                         & \WIKITROLD                & \REVISIONCOLOR 40.59/0.41                   & \REVISIONCOLOR 56.72/0.80                   & \REVISIONCOLOR 63.78/0.93                       & \REVISIONCOLOR 43.69/0.44                   & \REVISIONCOLOR 59.92/0.89          & \REVISIONCOLOR 67.56/1.35                   & \REVISIONCOLOR 52.10/0.52                   & \REVISIONCOLOR 70.67/1.03          & \REVISIONCOLOR 79.32/1.48                   \\
                                                                                            & \WIKITRNEW                & \REVISIONCOLOR 38.15/0.38                   & \REVISIONCOLOR 54.37/0.77                   & \REVISIONCOLOR 62.61/0.92                       & \REVISIONCOLOR 40.92/0.41                   & \REVISIONCOLOR 57.48/0.86          & \REVISIONCOLOR 66.22/1.32                   & \REVISIONCOLOR 48.99/0.49                   & \REVISIONCOLOR 68.23/0.99          & \REVISIONCOLOR 77.90/1.46                   \\
            \midrule

ColBERT-QA                                                            & \WIKITROLD                & \REVISIONCOLOR 58.99/0.59                   & \REVISIONCOLOR \textbf{70.34}/1.05          & \REVISIONCOLOR 72.77/1.38                       & \REVISIONCOLOR 62.27/0.62                   & \REVISIONCOLOR \textbf{74.03}/1.15 & \REVISIONCOLOR 78.07/1.64                   & \REVISIONCOLOR 75.88/0.76                   & \REVISIONCOLOR \textbf{88.23}/1.38 & \REVISIONCOLOR 92.10/1.83                   \\
                                                                                            & \WIKITRNEW                & \REVISIONCOLOR \textbf{60.50}/\textbf{0.61} & \REVISIONCOLOR \textbf{70.34}/\textbf{1.07} & \REVISIONCOLOR \textbf{74.54}/\textbf{1.44}     & \REVISIONCOLOR \textbf{63.78}/\textbf{0.64} & \REVISIONCOLOR 73.87/\textbf{1.18} & \REVISIONCOLOR \textbf{78.32}/\textbf{1.72} & \REVISIONCOLOR \textbf{77.05}/\textbf{0.77} & \REVISIONCOLOR 87.81/\textbf{1.40} & \REVISIONCOLOR \textbf{92.18}/\textbf{1.91} \\
            \bottomrule
        \end{tabular}%
        }
        \caption{Retriever results for the OpenQA formulation of QA task evaluated
        on \underline{\XQUADTR}. All tokenizers are uncased. The highest values in each
        column are shown in \textbf{bold}. For equal pairs, the larger ones on more
        significant digits are \underline{underlined}.}
        \label{tab:retrieval_results_xquad_tr}
    \end{table*}

    \begin{table*}
        [ht]
        \centering
        \setlength{\tabcolsep}{3.5pt}
        \resizebox{1\textwidth}{!}{%
        \begin{tabular}{>{\REVISIONCOLOR}l >{\REVISIONCOLOR}c >{\REVISIONCOLOR}c>{\REVISIONCOLOR}c>{\REVISIONCOLOR}c
        >{\REVISIONCOLOR}c>{\REVISIONCOLOR}c>{\REVISIONCOLOR}c >{\REVISIONCOLOR}c>{\REVISIONCOLOR}c>{\REVISIONCOLOR}c}
            \toprule                                                                        &                           & \multicolumn{3}{c}{\textbf{\REVISIONCOLOR Whitespace}} & \multicolumn{3}{c}{\textbf{\REVISIONCOLOR Morphological}} & \multicolumn{3}{c}{\textbf{\REVISIONCOLOR Enhanced Whitespace}} \\
            \cmidrule(l){3-5} \cmidrule(l){6-8} \cmidrule(l){9-11} \textbf{Retriever Model} & \textbf{Knowledge Source} & \textbf{S@1/\RCONE}                                    & \textbf{S@5/\RCFIVE}                                      & \textbf{S@20/\RCTWENTY}                                         %
                                                                                            & \textbf{S@1/\RCONE}       & \textbf{S@5/\RCFIVE}                                   & \textbf{S@20/\RCTWENTY}                                    %
                                                                                            & \textbf{S@1/\RCONE}       & \textbf{S@5/\RCFIVE}                                   & \textbf{S@20/\RCTWENTY}                                    \\
            \midrule BM25 - \emph{Baseline - Sparse}                                        & \WIKITROLD                & 19.04/\underline{0.19}                                 & 28.10/\underline{0.43}                                    & 35.09/0.73                                                     & 25.93/0.26                               & 37.38/0.60          & 45.43/1.02                   & 28.91/0.29                               & 40.53/\underline{0.65} & 48.39/1.07                   \\
                                                                                            & \WIKITRNEW                & 18.65/0.19                                             & 27.66/0.43                                                & 34.87/\underline{0.73}                                         & 25.42/0.25                               & 36.85/0.59          & 45.19/\underline{1.02}       & 28.31/0.28                               & 39.95/0.65             & 48.04/1.09                   \\
            \midrule

DPR - \emph{Baseline - Dense}                                         & \WIKITROLD                & 27.11/0.27                                             & 36.14/0.56                                                & 41.11/0.91                                                     & 37.18/\underline{0.37}                   & 47.70/0.78          & 52.47/1.29                   & 41.73/0.42                               & 51.91/0.86             & 56.01/1.37                   \\
                                                                                            & \WIKITRNEW                & 26.71/0.27                                             & 35.82/0.56                                                & 40.99/0.91                                                     & 36.64/0.37                               & 47.39/0.77          & 52.37/\underline{1.29}       & 41.17/0.41                               & 51.54/0.85             & 55.85/1.06                   \\
            \midrule

ColBERT-QA                                                            & \WIKITROLD                & \textbf{33.52}/\textbf{0.34}                           & \textbf{40.27}/\textbf{0.69}                              & 44.14/1.07                                                     & \textbf{46.07}/\underline{\textbf{0.46}} & \textbf{52.73}/0.95 & 56.22/1.50                   & \textbf{51.48}/\underline{\textbf{0.51}} & \textbf{56.97}/1.05    & 59.87/1.61                   \\
                                                                                            & \WIKITRNEW                & 33.18/0.33                                             & 40.25/\underline{\textbf{0.69}}                           & \textbf{44.40}/\textbf{1.10}                                   & 45.75/\textbf{0.46}                      & 52.72/\textbf{0.96} & \textbf{56.39}/\textbf{1.56} & 51.13/\textbf{0.51}                      & 56.82/\textbf{1.06}    & \textbf{59.91}/\textbf{1.67} \\
            \bottomrule
        \end{tabular}%
        }

        \caption{\REVISIONCOLOR Retriever results for the OpenQA formulation of QA
        task evaluated on \underline{\SQUADTRTRAIN}. All tokenizers are uncased. The
        highest values in each column are shown in \textbf{bold}. For equal pairs,
        the larger ones on more significant digits are \underline{underlined}.}
        \label{tab:retrieval_results_squad_tr_train}
    \end{table*}

    The {\REVISIONCOLOR\ performance} results of each retriever{\REVISIONCOLOR\ on \XQUADTR}
    are shown in Table~\ref{tab:retrieval_results_xquad_tr}. The results indicate that
    ColBERT~\citep{khattab2020colbert} is markedly more effective than the
    baseline BM25 and DPR models, even though the BM25 retriever is empowered with
    a morphological stemmer as described in \secref{sec:methods:openqa}. We observe
    a performance improvement over the BM25 and DPR~\citep{karpukhin-etal-2020-dense}
    models independent of the tokenization scheme. The results suggest that the \COLBERTQA
    retriever will give the reader module a better chance at finding correct answers.

    Comparing the baseline retriever models, we noticed a substantial performance
    advantage of BM25 over the DPR retriever. This finding is consistent with the reported
    performance of DPR~\citep{karpukhin-etal-2020-dense} on the English SQuAD 1.1~\citep{rajpurkar-etal-2016-squad}
    dataset. This is attributed to the fact that the annotators of the SQuAD
    datasets~\citep{rajpurkar-etal-2016-squad, rajpurkar-etal-2018-know} tended to
    formulate questions with significant lexical overlap with their passages,
    thereby providing an advantage to BM25. \citet{karpukhin-etal-2020-dense} also
    point out the skewed distribution of the target Wikipedia passages compared to
    the vast number of Wikipedia articles in the knowledge source as another
    contributing factor. {\REVISIONCOLOR\ This performance discrepancy also suggests that the DPR model is comparatively less effective in mitigating the inherent noise present in the knowledge source.}
    To address this issue, they suggested a hybrid approach that combines the
    outcomes of BM25 and DPR in order to achieve a result that surpasses each
    individually. {\REVISIONCOLOR\ Given that both models yield higher $S@k$ values as $k$ increases, these models can benefit from an effective reranker tailored to the noisy data for further improvement~\cite{du2022pregan}. }When
    we compare the baseline models with the ColBERT retriever, we observe that ColBERT
    achieves markedly superior performance than the baselines on{\REVISIONCOLOR\ \XQUADTR}.
    This indicates the outstanding effectiveness of ColBERT retriever in handling {\REVISIONCOLOR\ and ranking}
    examples characterized by lexical overlap as well as those requiring deep semantic
    understanding{\REVISIONCOLOR\, all while suppressing noise in the knowledge source and the training dataset}.

    The results also indicate that the morphology-unaware \emph{enhanced
    whitespace tokenizer} identifies the correct results better than the morphological
    tokenizer for all values of \emph{S@k} and \emph{C@k}, suggesting that
    computationally-intensive morphological stemming can be avoided when
    evaluating QA systems in Turkish. Although the negative effect of morphological
    stemming may be surprising given the rich morphology of Turkish, this result
    is in line with previous literature~\citep{budur-etal-2020-data}.

    {\REVISIONCOLOR\ As another perspective, we observed consistently diminishing results in the \emph{S@k} values in the BM25 and DPR models when utilizing the newer Turkish Wikipedia dump and evaluating on \XQUADTR. However, in the same scenario, we noted a consistent increase in most of the \emph{S@k} and \emph{C@k} scores for the \COLBERTQA model. It seems that, for DPR and BM25, the benefits of adding more relevant passages were outweighed by the interference effects from negative passages. In contrast, \COLBERTQA seems to be better able to suppress these interfering factors and benefit from the additional relevant data. }

    {\REVISIONCOLOR\
While the performance of the retriever models on \XQUADTR offers valuable insights, there is a need for a more effective method to explore how the retriever models interact with the reader model to improve the overall system performance. Given that the information transfer between the retrievers and readers primarily occurs through the retriever-specific training dataset prepared for the readers, we investigated the retrievers' performance on \SQUADTRTRAIN, as shown in Table~\ref{tab:retrieval_results_squad_tr_train}.

    The most notable observation we initially make in Table~\ref{tab:retrieval_results_squad_tr_train} compared to Table~\ref{tab:retrieval_results_xquad_tr} is the varying performance of DPR relative to BM25 on \XQUADTR and \SQUADTRTRAIN. This finding shows that DPR yields more examples with positive passages for the reader's training dataset compared to BM25. Consequently, we observe that DPR performs better on the machine-translated dataset compared to BM25, whereas BM25 surpasses DPR on the human-translated dataset. \COLBERTQA continues to demonstrate consistently strong results in both datasets.

    Another noticeable difference between Table~\ref{tab:retrieval_results_xquad_tr} and Table~\ref{tab:retrieval_results_squad_tr_train} is the slight decline in performance of the retriever models on \SQUADTRTRAIN as the knowledge source expands, evident across nearly all \emph{S@k} and \emph{C@k} scores for all models, including \COLBERTQA. This outcome suggests that the resulting training datasets for readers may contain fewer examples with positive passages, thus offering fewer chances to improve the respective readers. It is also worth noting that a decrease in the number of examples containing positive passages could affect reader scores differently depending on whether it occurs in the test dataset or the training dataset. Specifically, a reduction in the number of examples with positive passages in the test set may directly impact reader scores negatively. However, if the number of training examples is already sufficient to saturate the overall reader performance, a decrease in positive passages in the training dataset may have a lesser impact on reader scores. }

    We also evaluated the retrievers in terms of the ranking of the answers returned,
    which is a widely used metric in the information retrieval domain. In order to
    assess this property, we computed mean reciprocal rank (MRR) for the questions
    correctly answered by all retrievers by returning the same relevant passage
    among their top $k$ results at different ranks. All retrievers can return the same
    passage as the highest-ranked relevant passage among their top $k$ results,
    even though the passage's exact rank may differ in each retriever's output.{\REVISIONCOLOR\ Table~\ref{tab:retrieval_results_mrr_scores} shows the MRR scores of the retrievers on the examples where all retrievers return the same relevant passage with the highest rank among their top $k$ results. The scores are calculated on both \SQUADTRTRAIN and \XQUADTR with \WIKITROLD and \WIKITRNEW to reflect the ranking behaviors of the models on the training and test datasets as the knowledge source expands.}

    \begin{table*}
        [ht]
        \centering
        \setlength{\tabcolsep}{12pt}
        \begin{tabular}{>\REVISIONCOLOR l >\REVISIONCOLOR c >\REVISIONCOLOR c >\REVISIONCOLOR
        c >\REVISIONCOLOR c}
            \hline
                                                                          & \multicolumn{2}{c}{\textbf{\REVISIONCOLOR \WIKITROLD}} & \multicolumn{2}{c}{\textbf{\REVISIONCOLOR \WIKITRNEW}} \\
            \cmidrule(l){2-3} \cmidrule(l){4-5}

\textbf{Retriever Model} & \textbf{\SQUADTRTRAIN}                                 & \textbf{\XQUADTR}                                     & \textbf{\SQUADTRTRAIN} & \textbf{\XQUADTR} \\
            \midrule

BM25 - \emph{Baseline - Sparse}                     & 0.8262                                                 & 0.8879                                                & 0.8242                 & 0.8806            \\
            \midrule

DPR - \emph{Baseline - Dense}                       & 0.9267                                                 & 0.8474                                                & 0.9239                 & 0.8320            \\
            \midrule

\COLBERTQA                                          & 0.9867                                                 & 0.9575                                                & 0.9868                 & 0.9637            \\
            \bottomrule
        \end{tabular}
        \caption{The MRR scores of the retriever models evaluated on the ranks of
        the same relevant passages unanimously returned from all retrievers as the
        highest-ranked positive passages.}
        \label{tab:retrieval_results_mrr_scores}
    \end{table*}
    {\REVISIONCOLOR\
The MRR scores presented in Table~\ref{tab:retrieval_results_mrr_scores} offer two noteworthy insights. First, the DPR model consistently retrieves positive passages with higher confidence compared to BM25, regardless of the dataset or knowledge source used. While this behavioral difference is somewhat aligning with the relative performance of these retrievers on \SQUADTRTRAIN (Table~\ref{tab:retrieval_results_squad_tr_train}), it contradicts their relative performance on \XQUADTR (Table~\ref{tab:retrieval_results_xquad_tr}). This observation warrants further investigation when assessing the performance of the corresponding readers of these models. Second, the MRR scores of BM25 and DPR models decrease when the knowledge source expands, whereas the performance of \COLBERTQA improves. Once again, this finding suggests that \COLBERTQA can effectively leverage the potential increase in noise associated with the expansion of the knowledge source to better distinguish positive passages. }

    \paragraph{Qualitative Analysis of the Retriever}

    \begin{table*}
        []
        \resizebox{1\textwidth}{!}{%
        \begin{tabular}{l P{10cm} | P{10cm}}
            \toprule \toprule                                                    & \textbf{Turkish}                                                                                                                                                                                                                                                                                                                                                                                                                                                                                                                                                                                                                                                                                                                                                                                                                                              & \textbf{English}                                                                                                                                                                                                                                                                                                                                                                                                                                                                                                                                                                                                                                                                                                                                                                                                                   \\
            \midrule

\textbf{Question 165}                                      & Bir \underline{öğretmenlik} \underline{sertifikasının} geçerli olduğu en uzun süre nedir?                                                                                                                                                                                                                                                                                                                                                                                                                                                                                                                                                                                                                                                                                                                                                                     & What is the longest time that a \underline{teaching} \underline{certificate} is good for?                                                                                                                                                                                                                                                                                                                                                                                                                                                                                                                                                                                                                                                                                                                                          \\
            \midrule \textbf{Answer}                                             & on yıla                                                                                                                                                                                                                                                                                                                                                                                                                                                                                                                                                                                                                                                                                                                                                                                                                                                       & ten years                                                                                                                                                                                                                                                                                                                                                                                                                                                                                                                                                                                                                                                                                                                                                                                                                          \\
            \midrule

\textbf{BM25}                                              & ... için kullanılır. Genelde bu iptal bilgilerinin izlenmesinde kullanılır. Subject (Özne): \underline{Sertifikanın} ait olduğu varlık: bir cihaz, birey, ya da kurum. Issuer (Sağlayıcı): Bilgileri doğrulayan ve \underline{sertifikayı} imzalayan kuruluş. Not Before (Önce Değil): \underline{Sertifikanın} geçerli olduğu en erken saat ve tarihi. Not After (Sonra Değil): \underline{Sertifikanın} geçerli olduğu en geç saat ve tarihi. Key Usage (Anahtar Kullanımı): \underline{Sertifikanın} açık anahtarındaki geçerli kriptografik kullanım. Ortak alanlar arasında dijital imza doğrulaması, anahtar şifreleme ve \underline{sertifika} imzalama bulunur. Extended \ldots

\rightline{Source: \WIKITRNEW}                                                                                                                                       & It is used for monitoring the validity information. Subject: The entity to which the \underline{certificate} belongs: a device, individual, or organization. Issuer: The organization that verifies the information and signs the \underline{certificate}. Not Before: The earliest date and time at which the \underline{certificate} is valid. Not After: The latest date and time at which the \underline{certificate} is valid. Key Usage: The valid cryptographic usage in the public key of the \underline{certificate}. Common fields include digital signature verification, key encryption, and \underline{certificate} signing. Extended \ldots                                                                                                                                                                          \\
            \midrule \textbf{ColBERT-QA}                                         & \ldots yönelik gereksinimler, genelde tam zamanlı profesyonellere yönelik gereksinimler kadar sert değildir. İş Gücü İstatistikleri Bürosu, ABD'de 1,4 milyon ilkokul \underline{öğretmeni}, 674.000 ortaokul \underline{öğretmeni} ve 1 milyon lise \underline{öğretmeni} istihdam edildiğini tahmin etmektedir. Amerika Birleşik Devletleri'nde her eyalet devlet okullarında \underline{öğretmenlik} yapma lisansı almak için gereksinimleri belirler. \underline{Öğretim} \underline{sertifikasyonu} genelde üç yıl devam eder, ama \underline{öğretmenler} {\ANSWERCOLOR\textbf{on yıla}} varan uzunlukta \underline{sertifikalar} alabilirler. Devlet okulu \underline{öğretmenlerinin} bir lisans derecesine sahip olması şart koşulmakta ve \underline{öğretmenlerinin} çoğunun eğitim \ldots

\rightline{Source: \XQUADTR}                           & \ldots are generally not as rigorous as those for full-time professionals. The Bureau of Labor Statistics estimates that there are 1.4 million elementary school \underline{teachers}, 674,000 middle school \underline{teachers}, and 1 million secondary school \underline{teachers} employed in the U.S. In the United States, each state determines the requirements for getting a license to \underline{teach} in public schools. \underline{Teaching} \underline{certification} generally lasts three years, but \underline{teachers} can receive \underline{certificates} that last as long as {\ANSWERCOLOR\textbf{ten years}}. Public school \underline{teachers} are required to have a bachelor's degree and the majority \ldots                                                                                        \\
            \midrule \midrule

\textbf{Question 484}                             & \underline{Siliya} ne için kullanılır?                                                                                                                                                                                                                                                                                                                                                                                                                                                                                                                                                                                                                                                                                                                                                                                                                        & What are \underline{cilia} used for?                                                                                                                                                                                                                                                                                                                                                                                                                                                                                                                                                                                                                                                                                                                                                                                               \\
            \midrule \textbf{Answer}                                             & hareket yöntemi                                                                                                                                                                                                                                                                                                                                                                                                                                                                                                                                                                                                                                                                                                                                                                                                                                               & method of locomotion                                                                                                                                                                                                                                                                                                                                                                                                                                                                                                                                                                                                                                                                                                                                                                                                               \\
            \midrule \textbf{BM25}                                               & 1 milimetreden (0,039 in) 1,5 metreye (4,9 ft) kadar değişen boyutlarıyla taraklılar, ana {\ANSWERCOLOR\textbf{hareket yöntemi}} olarak \underline{siliya} (``kıl'') kullanan en büyük kolonyal olmayan hayvanlardır. Çoğu türün tarak dizisi denen ve vücutları boyunca devam eden, ktene adı verilen taraksı \underline{siliya} grupları taşıyan sekiz dizisi vardır ve böylece \underline{siliya} vurduğunda her bir tarak alttaki tarağa dokunur. ``Ktenofor'', Yunanca'da ``tarak'' anlamına gelen $\kappa\tau\epsilon\iota\zeta$                                                                                                                                                                                                                                                                                                                         
            (kök biçimi $\kappa\tau\epsilon\nu$-                                  
            ) ile ``taşıyan'' anlamına gelen Yunanca son ek -$\phi o\rho o \zeta$ 
            'tan gelir ve ``tarak taşıyan'' \ldots

\rightline{Source: \XQUADTR} & Ranging from about 1 millimeter (0.039 in) to 1.5 meters (4.9 ft) in size, ctenophores are the largest non-colonial animals that use \underline{cilia} (``hairs'') as their main {\ANSWERCOLOR\textbf{method of locomotion}}. Most species have eight strips, called comb rows, that run the length of their bodies and bear comb-like bands of \underline{cilia}, called ``ctenes'' stacked along the comb rows so that when the \underline{cilia} beat, those of each comb touch the comb below. The name ``ctenophora'' means ``comb-bearing'', from the Greek $\kappa\tau\epsilon\iota\zeta$                                                                                                                                                                                                                                                               
            (stem-form $\kappa\tau\epsilon\nu$-)                                  
            meaning ``comb'' and the Greek suffix -$\phi o\rho o \zeta$           
            meaning ``carrying''\ldots                                            \\
            \midrule \textbf{ColBERT-QA}                                         & \dashuline{Silikon} veya polisiloksan, siloksan'dan (-R2Si-O-SiR2-, burada R = organik grup) oluşan bir polimer'dir. Bunlar genellikle renksiz yağlar veya kauçuk benzeri maddelerdir. \dashuline{Silikonlar}, dolgu macunlarında, yapıştırıcılarda, yağlayıcılarda, tıpta, pişirme kaplarında, ısı ve elektrik yalıtımında kullanılır. Bazı yaygın biçimler arasında \dashuline{silikon} yağı, \dashuline{silikon} gresi, \dashuline{silikon} kauçuk, \dashuline{silikon} reçine ve \dashuline{silikon} kalafat bulunur. Daha kesin olarak polimerize edilmiş siloksan'lar veya polisiloksanlar olarak adlandırılan \dashuline{silikonlar}, her \dashuline{silikon} merkezine bağlı iki organik gruplu inorganik \dashuline{silikon}-oksijen omurga zinciri'nden ($\cdots$-Si-O-Si-O-Si-O-$\cdots$ oluşur. Genellikle \ldots

\rightline{Source: \WIKITRNEW} & \dashuline{Silicone} or polysiloxane is a polymer composed of \dashuline{siloxane} (-R2Si-O-SiR2-, where R = organic group). These are typically colorless oils or rubber-like substances. \dashuline{Silicones} are used in caulks, adhesives, lubricants, medicine, cooking utensils, and for thermal and electrical insulation. Some common forms of \dashuline{silicones} include \dashuline{silicone} oil, \dashuline{silicone} grease, silicone rubber, \dashuline{silicone} resin, and \dashuline{silicone} caulk. More precisely, \dashuline{silicones}, also called polymerized \dashuline{siloxanes} or polysiloxanes, consist of inorganic \dashuline{silicon}-oxygen backbone chains with two organic groups attached to each \dashuline{silicon} center ($\cdots$-Si-O-Si-O-Si-O-$\cdots$). They are generally \ldots \\
            \bottomrule \bottomrule
        \end{tabular}
        }
        \caption{The negative and positive effects of the TF-IDF approach used in
        BM25 are exemplified in Question~165 and Question~484, respectively. The
        content words in the questions and the corresponding correctly matched terms
        in the passages are shown underlined. The dashed lines represent the incorrectly
        matching terms (false positives) that adversely affect the results.}
        \label{tab:tfidf_ex}
    \end{table*}

    In order to observe the strengths and weaknesses of the retrievers relative to
    each other, we manually analyzed the passages retrieved for the questions in
    the test set by the two top-performing retrievers on \XQUADTR, \COLBERTQA~\citep{khattab-etal-2021-relevance}
    and BM25. The analysis revealed a number of factors that help to explain the performance
    differences between the sparse and dense retriever models~\citep{chen-etal-2022-salient}.

    One important factor is the TF-IDF-based scoring mechanism used in BM25, which
    results in the retrieval of irrelevant passages that excessively mention the uncommon
    content words in the questions. While this approach proves advantageous when
    there are only a few relevant candidate passages, it comes with significant side
    effects when the model needs to suppress multiple related passages for the question
    to return the actual relevant passage. The decrease in the \emph{S@k} and \emph{C@k}
    values for the BM25 retriever as the knowledge source expands (Table~\ref{tab:retrieval_results_xquad_tr})
    indicates that BM25 is ineffective in inhibiting new irrelevant signals. Table~\ref{tab:tfidf_ex}
    depicts two question--answer pairs\footnote{Question numbers are
    {\REVISIONCOLOR\ the sequence numbers} of the questions in \XQUADTR.} and passages
    returned by each retriever. The first example shows that TF-IDF-based scoring
    misleads the retriever and causes it to retrieve irrelevant passages
    containing the content words. Conversely, in the second example, where there were
    limited relevant candidate passages available, BM25 identified the correct
    passage. This observation aligns with the qualitative analysis conducted by \citet{karpukhin-etal-2020-dense}
    for English OpenQA, which compares the results of BM25 and DPR.

    \begin{table*}
        []
        \resizebox{1\textwidth}{!}{%
        \begin{tabular}{l P{10cm} | P{10cm}}
            \toprule \toprule                       & \textbf{Turkish}                                                                                                                                                                                                                                                                                                                                                                                                                                                                                                                                                                                                                                                                                                                                                                                                                                                                                                                                 & \textbf{English}                                                                                                                                                                                                                                                                                                                                                                                                                                                                                                                                                                                                                                                                                                                                                                                                                                                                                                                                              \\
            \midrule \textbf{Question 439}          & \underline{Amazon} \underline{Havzası}'nda kaç \underline{ülke} bulunmaktadır?                                                                                                                                                                                                                                                                                                                                                                                                                                                                                                                                                                                                                                                                                                                                                                                                                                                                   & How many \underline{nations} are within the \underline{Amazon} \underline{Basin}?                                                                                                                                                                                                                                                                                                                                                                                                                                                                                                                                                                                                                                                                                                                                                                                                                                                                             \\
            \midrule \textbf{Answer}                & dokuz                                                                                                                                                                                                                                                                                                                                                                                                                                                                                                                                                                                                                                                                                                                                                                                                                                                                                                                                            & nine                                                                                                                                                                                                                                                                                                                                                                                                                                                                                                                                                                                                                                                                                                                                                                                                                                                                                                                                                          \\
            \midrule

\textbf{BM25}                 & \underline{Amazon} \underline{Havzası}, Güney Amerika'nın \underline{Amazon} Nehri ve kolları tarafından beslenen bölümüdür. \underline{Amazon} drenaj \underline{havzası} 6.300.000 kilometrekare (2.400.000 sq mi) bir alanı kaplamaktadır ve bu değer Güney Amerika kıtasının yaklaşık \%35,5'ini oluşturmaktadır. \underline{Havza} Bolivya, Brezilya, Kolombiya, Ekvador, Fransız Guyanası (Fransa), Guyana, Peru, Surinam ve Venezuela \underline{ülkeleri} sınırları içinde yer almaktadır. \underline{Havzanın} çoğu \underline{Amazon} yağmur ormanları ile kaplıdır. Kapladığı 55 milyon kilometrekare ($21\times10^{6}$ sq mi) alan ile tropikal orman alanı, dünyanın en büyük yağmur ormanıdır. Dematteis, Lou; \ldots

\rightline{Source: \WIKITRNEW}                                                                                                                                                                              & The \underline{Amazon} \underline{basin} is the part of South America drained by the \underline{Amazon} River and its tributaries. The \underline{Amazon} drainage \underline{basin} covers an area of about 6,300,000 km2 (2,400,000 sq mi), or about 35.5 percent of the South American continent. It is located in the \underline{countries} of Bolivia, Brazil, Colombia, Ecuador, Guyana, Peru, Suriname, and Venezuela, as well as the territory of French Guiana.

Most of the \underline{basin} is covered by the \underline{Amazon} rainforest, also known as \underline{Amazonia}. With a 5.5 million km2 (2.1 million sq mi) area of dense tropical forest, it is the largest rainforest in the world. Dematteis, Lou; \ldots                                                                                                                                                                                                                      \\
            \midrule \textbf{ColBERT-QA}            & \underline{Amazon} yağmur ormanı (Portekizce: Floresta Amazônica veya Amazônia; İspanyolca: Selva \underline{Amazónica}, \underline{Amazonía} veya genellikle \underline{Amazonia}; Fransızca: Forêt \underline{amazonienne}; Hollandaca: \underline{Amazoneregenwoud}) İngilizce'de aynı zamanda \underline{Amazonia} veya \underline{Amazon} Jungle olarak da bilinir ve Güney Amerika'nın \underline{Amazon} \underline{havzasının} çoğunu kaplayan bir nemli geniş yapraklı ormanıdır. Bu \underline{havza} 7.000.000 kilometre karelik alanı kaplamaktadır (2.700.000 mil kare) ve bunun 5.500.000 kilometre karesi (2.100.000 mil kare) yağmur ormanıyla kaplıdır. Bu bölge {\ANSWERCOLOR\textbf{dokuz}} \underline{ulusa} ait toprakları içermektedir. Ormanın çoğu yağmur ormanının \%60'ı \ldots

\rightline{Source: \XQUADTR}                                                                                                          & The \underline{Amazon} rainforest (Portuguese: Floresta \underline{Amaz\^{o}nica} or \underline{Amaz\^{o}nia}; Spanish: Selva \underline{Amaz\'onica}, \underline{Amazon\'ia} or usually \underline{Amazonia}; French: Forêt \underline{amazonienne}; Dutch: \underline{Amazoneregenwoud}), also known in English as \underline{Amazonia} or the \underline{Amazon} Jungle, is a moist broadleaf forest that covers most of the \underline{Amazon} \underline{basin} of South America. This \underline{basin} encompasses 7,000,000 square kilometres (2,700,000 sq mi), of which 5,500,000 square kilometres (2,100,000 sq mi) are covered by the rainforest. This region includes territory belonging to {\ANSWERCOLOR\textbf{nine}} \underline{nations}. The majority of the forest is contained within Brazil, with 60\% \ldots                                                                                                                           \\
            \midrule \midrule \textbf{Question 922} & Hangi \underline{Nobel} \underline{Ekonomi} \underline{Ödülü} \underline{kazananı} aynı zamanda bir \underline{üniversite} \underline{mezun} \underline{üyesidir}?                                                                                                                                                                                                                                                                                                                                                                                                                                                                                                                                                                                                                                                                                                                                                                               & What \underline{Nobel} \underline{Memorial} \underline{Prize} in \underline{Economic} \underline{Sciences} \underline{winner} is also a \underline{university} \underline{alumni} \underline{member}?                                                                                                                                                                                                                                                                                                                                                                                                                                                                                                                                                                                                                                                                                                                                                         \\
            \midrule \textbf{Answer}                & Milton Friedman                                                                                                                                                                                                                                                                                                                                                                                                                                                                                                                                                                                                                                                                                                                                                                                                                                                                                                                                  & Milton Friedman                                                                                                                                                                                                                                                                                                                                                                                                                                                                                                                                                                                                                                                                                                                                                                                                                                                                                                                                               \\
            \midrule

\textbf{BM25}                 & \underline{Üniversitelerine} göre \underline{Nobel} \underline{Ödülü} sahipleri listesi, \underline{Nobel} \underline{Ödülü} \underline{kazananların} (öğrenci veya \underline{mezun} oldukları \underline{üniversitelere} göre) birinci derecede eğitim gördükleri \underline{Üniversitelere} göre listelenmiş halidir. \underline{Üniversiteler} \underline{Nobel} \underline{Ödülü} \underline{kazananların} sayısına göre doğru orantılı şekilde sıralanmıştır. \underline{Üniversitelere} göre listeleme işlemi oldukça kapsamlı bir çalışma gerektirmiştir. Bu nedenle çok çeşitli kaynaklardan yararlanılmıştır. \underline{Ödülü} \underline{kazanan} birçok kişi farklı \underline{Üniversitelere} geçiş veya doktora yapmıştır. Bu nedenle bazı \underline{ödül} sahipleri birden fazla \underline{Üniversitede} eğitim görmüş veya doktora yapmış olabilir. Aşağıdaki listede \underline{ödül} \ldots

\rightline{Source: \WIKITRNEW} & The list of \underline{Nobel} \underline{Prize} \underline{laureates} according to their \underline{universities} is a compilation that categorizes the \underline{winners} (based on whether they were students or graduates) according to the \underline{universities} where they received their primary education. The \underline{universities} are ranked in proportion to the number of \underline{Nobel} \underline{Prize} recipients. The process of listing the \underline{universities} required extensive and comprehensive research, utilizing various sources. Many \underline{prize} \underline{winners} have transitioned to different \underline{universities} or pursued doctoral degrees, resulting in some recipients having received education or completed their doctorates at multiple \underline{universities}. The list below features the recipients of the award.                                                                    \\
            \midrule \textbf{ColBERT-QA}            & \ldots yer alır. Amerikalı ekonomist, sosyal kuramcı, politif filozof ve yazar Thomas Sowell de \underline{üniversitenin} \underline{mezunları} arasındadır. \underline{Ekonomide}, tanınmış \underline{Nobel} \underline{Ekonomi} \underline{Ödüllü}, ABD'nin Cumhuriyetçi Başkanı Ronald Reagan'ın Muhafazakar Britanya Başbakanı Margaret Thatcher'ın baş danışmanlarından biri olan {\ANSWERCOLOR\textbf{Milton Friedman}}, \underline{Nobel} \underline{ödüllü} ve düzenleme tuzağı teorisini ileri süren George Stigler, \underline{ekonominin} aile \underline{ekonomisi} dalına önemli katkılar sunmuş olan Gary Becker, örgütsel karar verme konseptinin modern yorumundan sorumlu Herbert A. Simon, \underline{Nobel} \underline{Ekonomi} \underline{Ödüllü} ilk Amerikalı olan Paul Samuelson \ldots

\rightline{Source: \XQUADTR}                                                                                                    & \ldots are. American \underline{economist}, social theorist, political philosopher, and author Thomas Sowell is also an \underline{alumnus}. In \underline{economics}, notable \underline{Nobel} \underline{Memorial} \underline{Prize} in \underline{Economic} \underline{Sciences} \underline{winners} {\ANSWERCOLOR\textbf{Milton Friedman}}, a major advisor to Republican U.S. President Ronald Reagan and Conservative British Prime Minister Margaret Thatcher, George Stigler, \underline{Nobel} laureate and proponent of regulatory capture theory, Gary Becker, an important contributor to the family \underline{economics} branch of \underline{economics}, Herbert A. Simon, responsible for the modern interpretation of the concept of organizational decision-making, Paul Samuelson, the first American to \underline{win} the \underline{Nobel} \underline{Memorial} \underline{Prize} in \underline{Economic} \underline{Sciences} \ldots \\
            \bottomrule \bottomrule
        \end{tabular}
        }
        \caption{Examples showcasing how ColBERT-QA can better capture the
        information needs implicit in questions. The content words in the questions and
        the corresponding correctly matched terms in the passages are shown underlined.}
        \label{tab:detailed_info_ex}
    \end{table*}

    Another factor is the ability of \COLBERTQA to represent questions better than
    BM25 and thus retrieve relevant passages more accurately. Two example questions
    are given in Table~\ref{tab:detailed_info_ex}. In the first one, both models retrieved
    passages related to the Amazon (region) and its surrounding countries. However,
    only the passage retrieved by \COLBERTQA provided a specific numerical answer to
    the question ``how many''. In contrast, the passage retrieved by BM25 consisted
    of the correct list of the countries without an explicit count, posing a
    challenge for the reader in extracting the correct answer span. In the second example,
    \COLBERTQA successfully identified the information need as a Nobel Prize
    winner who is also a member of a university alumni, and retrieved the relevant
    passage. In contrast, BM25 retrieved a generic passage about universities and the
    Nobel Prize, missing the specific person targeted as the information need.

    \begin{table*}
        []
        \resizebox{1\textwidth}{!}{%
        \begin{tabular}{l P{10cm}|P{10cm}}
            \toprule \toprule                        & \textbf{Turkish}                                                                                                                                                                                                                                                                                                                                                                                                                                                                                                                                                                                                                                                                                                                                                   & \textbf{English}                                                                                                                                                                                                                                                                                                                                                                                                                                                                                                                                                                                                                                                                                                                           \\
            \midrule

\textbf{Question 430}          & \underline{Kaç} \underline{ülke} \underline{isminde} \underline{``Amazonas''} \underline{bulunmaktadır}?                                                                                                                                                                                                                                                                                                                                                                                                                                                                                                                                                                                                                                                           & \underline{How many} \underline{nations} \underline{contain} \underline{``Amazonas''} in their \underline{names}?                                                                                                                                                                                                                                                                                                                                                                                                                                                                                                                                                                                                                          \\
            \midrule \textbf{Answer}                 & Dört                                                                                                                                                                                                                                                                                                                                                                                                                                                                                                                                                                                                                                                                                                                                                               & four                                                                                                                                                                                                                                                                                                                                                                                                                                                                                                                                                                                                                                                                                                                                       \\
            \midrule \textbf{BM25}                   & Tarık el-Tayyib Muhammed Buazizi (29 Mart 1984 - 4 Ocak 2011), Tunuslu seyyar satıcı. 17 Aralık 2010'da kendisini yakarak intihar girişiminde \dashuline{bulundu}. Bu olayın tesiri ile Tunus halkının ayaklanması üzerine 23 yıldır \underline{ülkeyi} yöneten Zeynel Abidin Bin Ali \underline{ülkeden} \dashuline{kaçmıştır}. Bu olay aynı zamanda diğer Arap ülkelerindeki ayaklanmaları teşvik etmiştir. Ölümünden sonra Tunus'ta Yasemin Devrimi başlamıştır. 17 Ocak 2011'de başkent Tunus'un en ünlü caddesi olan 7 Kasım Caddesi'nin \underline{ismi} (Zeynel \ldots

\rightline{Source: \WIKITRNEW}                                                                                                                                                      & Tarık el-Tayyib Muhammed Buazizi (March 29, 1984 - January 4, 2011) was a Tunisian street vendor. On December 17, 2010, he \dashuline{set himself on} fire in a suicide attempt. As a result of this incident, Zine El Abidine Ben Ali, who had been ruling the \underline{country} for 23 years, \dashuline{fled} from Tunisia. This event also inspired uprisings in other Arab countries. Following his death, the Jasmine Revolution began in Tunisia. On January 17, 2011, the \underline{name} of 7 November Avenue, the most famous street in the capital city of Tunis (Zine El \ldots                                                                                                                                             \\
            \midrule                                  
            \textbf{ColBERT-QA}                      & \ldots ile Brezilya sınırları içindedir, ardından \%13 ile Peru, \%10 ile Kolombiya, ve daha az oranlarla Venezuela, Ekvador, Bolivya, Guyana, Surinam ve Fransız Guyanası gelir. {\ANSWERCOLOR\textbf{Dört}} \underline{ülkenin} eyalet veya il \underline{isimlerinde} \underline{``Amazonas''} geçmektedir. \underline{Amazon} gezegenin mevcut yağmur ormanlarının yarıdan fazlasını temsil etmektedir ve dünyadaki en büyük ve en çok biyoçeşitliliğe sahip tropik yağmur ormanı alanını \underline{içermektedir} ve buna 16.000 türe ayrılan 390 milyar ağaç dahildir. \underline{Amazon} yağmur ormanı (Portekizce: Floresta \underline{Amazônica} veya \underline{Amazônia}; İspanyolca: Selva \underline{Amazónica}, \ldots

\rightline{Source: \XQUADTR} & \ldots is contained within Brazil, followed by Peru with 13\%, Colombia with 10\%, and with minor amounts in Venezuela, Ecuador, Bolivia, Guyana, Suriname and French Guiana. States or departments in {\ANSWERCOLOR\textbf{four}} \underline{nations} \underline{contain} \underline{``Amazonas''} in their \underline{names}. The \underline{Amazon} represents over half of the planet's remaining rainforests, and comprises the largest and most biodiverse tract of tropical rainforest in the world, with an estimated 390 billion individual trees divided into 16,000 species. The \underline{Amazon} rainforest (Portuguese: Floresta \underline{Amazônica} or \underline{Amazônia}; Spanish: Selva \underline{Amazónica} \ldots \\
            \midrule \midrule

\textbf{Question 941} & \underline{Huihui} neydi?                                                                                                                                                                                                                                                                                                                                                                                                                                                                                                                                                                                                                                                                                                                                          & What was \underline{huihui}?                                                                                                                                                                                                                                                                                                                                                                                                                                                                                                                                                                                                                                                                                                               \\
            \midrule \textbf{Answer}                 & Müslüman tıbbı                                                                                                                                                                                                                                                                                                                                                                                                                                                                                                                                                                                                                                                                                                                                                     & Muslim medicine                                                                                                                                                                                                                                                                                                                                                                                                                                                                                                                                                                                                                                                                                                                            \\
            \midrule

\textbf{BM25}                  & Batı tıbbı, bazen \underline{huihui} ya da {\ANSWERCOLOR\textbf{Müslüman tıbbı}} olarak adlandırıldığı Yuan meclisinin Nestûrî Hristiyanları tarafından Çin'de de uygulanmıştır. Nestûrî hekim Tercüman İsa, 1963 yılında, Kubilay'ın saltanatı döneminde Batı Tıbbı Ofisini kurmuştur. İki imparatorluk hastanesinde çalışan doktorlar imparatorluk ailesi ve meclisin üyelerini tedavi etmekten sorumluydu. Çinli hekimler, hümoral sistemi, geleneksel Çin tıbbının altında yatan yin-yang ve wuxing felsefesine karşı geldiği için Batı tıbbına karşı çıkıyorlardı. Batı tıbbı çalışmalarının bilinen bir Çin tercümesi yoktur ama Çinlilerin İbn-i \ldots

\rightline{Source: \XQUADTR}                                                                       & Western medicine was also practiced in China by the Nestorian Christians of the Yuan court, where it was sometimes labeled as \underline{huihui} or {\ANSWERCOLOR\textbf{Muslim medicine}}. The Nestorian physician Jesus the Interpreter founded the Office of Western Medicine in 1263 during the reign of Kublai. \underline{Huihui} doctors staffed at two imperial hospitals were responsible for treating the imperial family and members of the court. Chinese physicians opposed Western medicine because its humoral system contradicted the yin-yang and wuxing philosophy underlying traditional Chinese medicine. No Chinese translation of Western medical works is known, but Chinese had Avicenna \ldots                    \\
            \midrule \textbf{ColBERT-QA}             & Tüzükleri, işbirliği yapmayan çocuk işçiler için hapis şartlarını öngörmüştür. Hong Kong gibi güneydoğu Asya kolonilerinde \dashuline{Mui} Tsai () gibi çocuk işçiliği kültürel bir gelenek olarak rasyonelleştirildi ve İngiliz yetkililer tarafından göz ardı edildi. Hollanda Doğu Hindistan Şirketi yetkilileri, çocuklarının işçi tacizlerini “bu, bu çocukları daha kötü bir kaderden kurtarmanın bir yolu” ile mantıklı hale getirdiler. Zambiya'dan Nijerya'ya uzanan bölgelerdeki Hıristiyan misyon okulları da çocuklardan çalışma gerektirdi ve karşılığında laik eğitim değil din eğitimi sağladı. Başka \ldots

\rightline{Source: \SQUADTRTRAIN}                                                                                                     & In southeast Asian colonies, such as Hong Kong, child labour such as the \dashuline{Mui} Tsai (), was rationalised as a cultural tradition and ignored by British authorities. The Dutch East India Company officials rationalised their child labour abuses with, ``it is a way to save these children from a worse fate.'' Christian mission schools in regions stretching from Zambia to Nigeria too required work from children, and in exchange provided religious education, not secular education. Elsewhere \ldots                                                                                                                                                                                                                 \\
            \bottomrule \bottomrule
        \end{tabular}
        }
        \caption{Examples that illustrate how WordPiece tokenization can produce a
        mix of favorable and unfavorable outcomes, depending on its ability to resist
        the influence of lexical bias. The content words in the questions and the
        corresponding correctly matched terms in the passages are shown underlined. The
        dashed lines represent the incorrectly matching terms (false positives) that
        adversely affect the results.}
        \label{tab:wordpiece_ex}
    \end{table*}
    During manual analysis, we also observed an intriguing aspect related to \COLBERTQA's
    WordPiece tokenization, which can have both positive and negative implications.
    Table~\ref{tab:wordpiece_ex} shows two example cases. In the first example, \COLBERTQA
    employed WordPiece tokenization to split the word ``Amazonas'' into the word pieces
    \mbox{\texttt{[``Amazon'', "\#\#as"]}}. This split allowed \COLBERTQA to
    correctly associate the word ``Amazonas'' with the word ``Amazon'' and
    successfully retrieve the relevant passage. On the other hand, BM25 placed
    excessive emphasis on the term ``Amazonas'' and other content words in the
    question due to its lexical bias, leading to the retrieval of an entirely unrelated
    passage that contained these content words extensively.

    However, WordPiece tokenization can be a liability as well. In the second example,
    despite the word ``Huihui'' being a proper noun, \COLBERTQA tokenized it as \mbox{\texttt{[``Hu'',
    ``\#\#ih'', ``\#\#u'', ``\#\#i'']}}, resulting in retrieving an irrelevant
    passage. The same effect can also be seen in Question 484 in Table~\ref{tab:tfidf_ex},
    where \COLBERTQA matched the words
    \mbox{``Silikon'' \texttt{[``Sili'', ``\#\#kon'']}} and
    \mbox{``Siliya'' \texttt{[``Sili'', ``\#\#ya'']}} due to their common prefix
    and incorrectly retrieved a passage on ``Silikon''~(Silicone) for a question
    about ``Siliya'' (Cilia), which are completely different concepts.

    \begin{table*}
        []
        \resizebox{1\textwidth}{!}{%
        \begin{tabular}{>\REVISIONCOLOR l >\REVISIONCOLOR P{10cm} | >\REVISIONCOLOR
        P{10cm}}
            \toprule \toprule               & \textbf{Turkish}                                                                                                                                                                                                                                                                                                                                                                                                                                                                                                                                                                                                                                                                                                                                                                                                                                                                                                                                                                                                                                                                                                                          & \textbf{English}                                                                                                                                                                                                                                                                                                                                                                                                                                                                                                                                                                                                                                                                                                                                                                                                                                                                                                                                                                                        \\
            \midrule

\textbf{Question 178} & \underline{2009}'da \underline{Beyonce} \underline{ikinci} \underline{dünya} \underline{turuna} \underline{başladı} ve \underline{ne kadar} \underline{hasılat} \underline{etti}?                                                                                                                                                                                                                                                                                                                                                                                                                                                                                                                                                                                                                                                                                                                                                                                                                                                                                                                                                         & In \underline{2009}, \underline{Beyonce} \underline{started} her \underline{second} \underline{world} \underline{tour} and \underline{grossed} \underline{how much} money?                                                                                                                                                                                                                                                                                                                                                                                                                                                                                                                                                                                                                                                                                                                                                                                                                              \\
            \midrule \textbf{Answer}        & 119,5 milyon                                                                                                                                                                                                                                                                                                                                                                                                                                                                                                                                                                                                                                                                                                                                                                                                                                                                                                                                                                                                                                                                                                                              & 119.5 million                                                                                                                                                                                                                                                                                                                                                                                                                                                                                                                                                                                                                                                                                                                                                                                                                                                                                                                                                                                           \\
            \midrule

\textbf{BM25}         & The \underline{Beyoncé} Experience, Amerikalı şarkıcı \underline{Beyoncé}'nin üçüncü konser turnesi. \underline{Beyoncé}, 4 Eylül 2006'da \underline{ikinci} solo albümü B'Day'i yayımlamıştı. Albümün getirdiği başarı onu ilk \underline{dünya} turnesine götürdü. Turnenin adının ``B'Day World Tour'' olması düşünülmüştü ancak \underline{Beyonce}'nin ilk \underline{dünya} \underline{turu} deneyimi olduğu için ``The \underline{Beyonce} Experience Tour\'' olmasına karar verildi. Sony Music stüdyolarında gerekli çalışmalar, provalar, dansçı seçimleri, koreografiler, sahne dekoru ayarlamaları yapıldıktan sonra Mart'ta genel hazırlıklar \underline{başladı}. Normalde Avrupa'da başlayacak olan turne bazı aksaklıklar yüzünden Japonya'dan \ldots

\rightline{Source: \WIKITRNEW}                                                                                                                                                                                                                                                                                                                                     & The \underline{Beyoncé} Experience is the third concert tour by American singer \underline{Beyoncé}. \underline{Beyoncé} had released her \underline{second} solo album, B'Day, on September 4, 2006. The success of the album led her to embark on her first \underline{world} tour. The tour was initially planned to be called the ``B'Day World Tour,'' but it was decided to name it ``The \underline{Beyonce} Experience Tour'' as it was \underline{Beyoncé}'s first experience with a \underline{world} \underline{tour}. After necessary work, rehearsals, dancer selections, choreography, and stage set adjustments were made at Sony Music studios, general preparations \underline{began} in March. The tour, which was originally planned to start in Europe, (began) from Japan due to some setbacks. \ldots                                                                                                                                                                             \\
            \midrule \textbf{DPR}           & \ldots Kadın Video kategorisini \underline{kazanamaması}, Kanye West'in töreni kesintiye uğratmasına ve \underline{Beyoncé}'nin kendi kabul konuşması sırasında Swift'in ödülünü yeniden sunumunu gerçekleştirmesine yol açtı. Mart 2009'da, \underline{Beyoncé} I... \underline{Dünya} \underline{Turu}, 108 gösteriden oluşan \underline{dünya} çapında konser \underline{turu}, {\ANSWERCOLORINREVISION \textbf{119,5 milyon}} dolar \underline{hasılat} kazanıyor. 4 Nisan 2008'de, Beyoncé Jay Z ile evlendi. Üçüncü stüdyo albümü olan I Am'in dinleme partisinde bir video montajında evliliklerini açıkça açıkladı. Sasha Fierce, 22 Ekim 2008'de Manhattan'ın Sony Kulübünde. Ben... Sasha Fierce 18 Kasım \ldots

\rightline{Source: \SQUADTRTRAIN}                                                                                                                                                                                                                                                                                                                                                                             & \ldots Not \underline{winning} in the Female Video category led to Kanye West interrupting the ceremony and \underline{Beyoncé} re-presenting Swift's award during her acceptance speech. In March 2009, \underline{Beyoncé}'s I... \underline{World} \underline{Tour}, a \underline{world}wide concert tour consisting of 108 shows, \underline{gross}ing {\ANSWERCOLORINREVISION\textbf{\$119.5 million}} in revenue. On April 4, 2008, \underline{Beyoncé} married Jay Z. They openly announced their marriage in a video montage at the listening party for her third studio album, I Am... Sasha Fierce, on October 22, 2008, at Manhattan's Sony Club. I... Sasha Fierce November 18th. \ldots                                                                                                                                                                                                                                                                                                    \\
            \midrule

\textbf{\COLBERTQA}   & \ldots Amerikalı şarkıcı Taylor Swift'in ``You Belong with Me'' şarkısının klibine giden En İyi Kadın Klibi kategorisinde ödül alamaması, Kanye West'in töreni durdurmasına ve Beyoncé'nin kendi ödül konuşmasını Swift'e vermesine yol açtı. Mart $\underset{\text{\relsize{2} 2009}}{\underline{2009}}$'da $\underset{\text{\relsize{2} Beyonce}}{\underline{Beyonce}}$, 108 gösteriden oluşan ve {\ANSWERCOLORINREVISION\$\textbf{119,5}} \textbf{$\underset{\textnormal{\relsize{2} ne kadar}}{\underline{\textbf{\ANSWERCOLORINREVISION milyon}}}$} $\underset{\text{\relsize{2} hasılat}}{\underline{hasilat}}$ $\underset{\text{\relsize{2} etti}}{\underline{yapan}}$ $\underset{\text{\relsize{2} ikinci}}{\underline{ikinci}}$ $\underset{\text{\relsize{2} dünya}}{\underline{dunya}}$ turnesi I Am... World $\underset{\text{\relsize{2} turuna}}{\underline{Tour}}$'u başlattı. Beyoncé, 2008 yapımı müzikal biyografik film Cadillac Records'ta blues şarkıcısı Etta James olarak başrolde yer alarak filmlerde rol almaya devam etti. Filmdeki performansı eleştirmenler tarafından \ldots

\rightline{Source: \WIKITRNEW} & \ldots The failure of American singer Taylor Swift to win the Best Female Video category for her song ``You Belong with Me'' led to Kanye West interrupting the ceremony and Beyoncé giving her own award speech to Swift. In March $\underset{\text{\relsize{2} 2009}}{\underline{2009}}$, $\underset{\text{\relsize{2} Beyonce}}{\underline{Beyonce}}$ launched her $\underset{\text{\relsize{2} second}}{\underline{second}}$ $\underset{\text{\relsize{2} world}}{\underline{world}}$ $\underset{\text{\relsize{2} tour}}{\underline{tour}}$, I Am... World tour, consisting of 108 shows and $\underset{\text{\relsize{2} grossed}}{\underline{grossing}}~${\ANSWERCOLORINREVISION\$\textbf{119.5}} $\underset{\textnormal{\relsize{2} how much}}{\underline{\textbf{\ANSWERCOLORINREVISION million}}}$. Beyoncé continued to act in films, starring as blues singer Etta James in the 2008 musical biographical film Cadillac Records. Her performance in the film was praised by critics. \ldots \\
            \bottomrule \bottomrule
        \end{tabular}
        }
        \caption{\REVISIONCOLOR Example passages chosen by each retriever for a
        given question from \SQUADTRTRAIN, and put into the training set of the corresponding
        readers.}
        \label{tab:example_for_all_retriever_results}
    \end{table*}

    { \REVISIONCOLOR\
 To further deepen this analysis, Table~\ref{tab:example_for_all_retriever_results} shows sample passages selected by each retriever for a particular question sourced from \SQUADTRTRAIN. These passages are incorporated into the training set of the respective readers.

    In this particular example, we see that BM25 fails to identify the relevant passage containing the answer for the question due to its lexical nature: the incorrect passage it selected contains numerous repetitions of one of the key content words in the query.

    To analyze DPR and \COLBERTQA, we used BertViz~\cite{vig-2019-multiscale} as a tool to visualize the output of \mbox{BERT-based} models.\footnote{\REVISIONCOLOR\ The visual representations are omitted from the paper due to concerns about their readability within the constraints of paper size. However, we provide descriptions of the key insights from the visualizations in the text and make the full-sized visualizations on our Github page: \SQUADTRURLPROD} When DPR encodes the question in Table~\ref{tab:example_for_all_retriever_results}, we see that the \texttt{[CLS]} token in the output layer is attending to the \texttt{[CLS]} token in the preceeding layer. This behavior continues across the other layers until the earlier layers, where the multi-head attention is distributed across all the words but focuses relatively more on the content words \textit{Beyoncé, ne kadar (how much), hasılat etti (grossing)} in certain heads. When DPR encodes the passage, a similar pattern is observed for the \texttt{[CLS]} token, where the attention of the \texttt{[CLS]} token in the earlier layers is spread across all tokens in the passage with loose emphasis on certain content words like \textit{Beyoncé, dünya turu (world tour), hasılat (gross), as underlined in the passage for DPR}.

    For \COLBERTQA, we also show in the table the pairs of question-passage tokens matched by means of MaxSim scoring in the passage. \COLBERTQA establishes a balance between lexical matching and semantic matching. For example, it is able to match certain question words such as \textit{2009, Beyoncé, hasılat (gross), ikinci (second), dünya~(world)} with the corresponding words in the paragraph in an exact manner, just like BM25 does. On the other hand, it can also match certain words and phrases such as \textit{ne kadar (how much)} and \textit{etti} (akin to \textit{made}) with the semantically related words and phrases \textit{milyon (million)} and \textit{yapan} (akin to \textit{doing}), respectively, demonstrating semantic association. Unlike BM25, \COLBERTQA is unaffected by the excessive presence of a question token in the passage, as it only selects one passage token that is most similar to the question token. This observation suggests a potential avenue for future research: modifying \COLBERTQA to allow the selection of multiple passage tokens for each question token, thereby viewing each scoring as a mini ranking function that operates on the tokens within a passage given a question token. We leave this exploration for a future work.

    The example presented in Table~\ref{tab:example_for_all_retriever_results} also demonstrates how the retriever models function in scenarios where potential noise is introduced by machine translation. In the question, the phrase \emph{``hasılat etti''} is a flawed translation used in lieu of the correct translation \emph{``hasılat yaptı''} for the English term \emph{``grossed''}. This discrepancy arises from the translation system's confusion between the words \emph{``etti''} (akin to \emph{``made''}) and \emph{``yaptı''} (akin to \emph{``did''}) owing to the difference between the use of these words in Turkish and English. For this question, DPR returns the target positive passage sourced from \SQUADTRTRAIN, which is a machine-translated text, while \COLBERTQA retrieves the equivalent passage from Wikipedia. BM25 returns a negative passage from Wikipedia due to its lexical term bias.

    The passage returned by DPR has the expression \emph{``Beyoncé I\ldots Dünya Turu''} (\emph{``Beyoncé I\ldots World Tour''}), which is an erroneous translation. The correct translation for the original expression \emph{``Beyoncé I Am\ldots World Tour''} should have been \emph{``Beyoncé I Am\ldots Dünya Turu''}, as seen in the corresponding Wikipedia passage retrieved by \COLBERTQA. In this particular example, we understand that each model handles the noise in the translated texts in a different way. BM25 completely disregards the translation error in the question term because it does not find a matching term in the retrieved passage. DPR successfully retrieves the target passage from \SQUADTRTRAIN as its fixed-size representations align with those of the question, without being affected by the translation error that exists in both the question and the passage. Meanwhile, \COLBERTQA retrieves the original Wikipedia passage by associating each question token with the most semantically relevant passage token, even in cases where the question token was inaccurately translated. For instance, the question token \textit{etti} (akin to \textit{made}) was matched with the closest semantically related passage token \textit{yapan} (akin to \textit{doing}). In this respect, \COLBERTQA surpasses DPR in the quality of passages returned, by effectively managing the individual interactions between question and passage tokens, as seen in BM25. Furthermore, it also outperforms BM25 by leveraging its distributional representation capability to match semantically related tokens. }

    \subsubsection{Reader Results}

    Table~\ref{tab:openqa_reader_results} shows the results of the reader step of
    the OpenQA formulation. The results demonstrate that the reader trained on the
    dataset obtained by the \COLBERTQA~\citep{khattab-etal-2021-relevance}
    retriever using \WIKITROLD achieves around
    {\REVISIONCOLOR\ 27\%~(EM) / 23\%~(F1)} improvement and around
    {\REVISIONCOLOR\ 26\%~(EM) / 22\%~(F1)} improvement compared to the readers
    that use the baseline{\REVISIONCOLOR\ BM25 and DPR retrievers, respectively}.
    {\REVISIONCOLOR\ This improvement changes to around 24\%~(EM) / 26\%~(F1) for BM25 and to approximately 33\%~(EM) / 29\%~(F1) for DPR when the retrievers use \WIKITRNEW.}
    {\REVISIONCOLOR\ Based on these findings, we hypothesize that the substantial degradation in the quality of the bootstrap training datasets for BM25 and DPR relative to \COLBERTQA, as the knowledge source grows over time, is the source of the increasing gap between the baseline models and \COLBERTQA. Similarly, the higher quality training dataset generated by \COLBERTQA for its reader, as the knowledge-source expands, contributes to narrowing the gap between the \COLBERTQA-based reader towards the upper bound standard QA reader results shown in Table \ref{tab:standard_formulation_reader_results}. This is a striking finding that underscores the capacity of the retriever models to withstand the gradual expansion of the knowledge source over time.}

    {\REVISIONCOLOR\  To further explore the effects of expanding the knowledge source, we inspected the reader outputs in relation to the source of the passages. Our aim in this analysis is to determine if the number of passages retrieved from the knowledge source increases or not as the knowledge source expands, and how this impacts the performance of the readers.

    Table~\ref{tab:openqa_reader_results_error_analysis} shows the outputs of each reader categorized by the source of passages selected from \WIKITROLD and \WIKITRNEW. The BM25-based reader includes about 26\% of passages from Wikipedia in their test datasets when using \WIKITROLD\ and 27\% when using \WIKITRNEW. For \DPRBASED\ readers, these numbers are 20\% for \WIKITROLD\  and 22\% for \WIKITRNEW. However, for \COLBERTQA, these numbers are much lower: 4\% from \WIKITROLD and 6\% from \WIKITRNEW. When passages are from Wikipedia, the success rates of these readers are somewhat variable. For the BM25-based readers, the success rate is around 6\% for both \WIKITROLD\ and \WIKITRNEW. For \DPRBASED\ readers, these numbers are 5\% and 3\%, respectively. For \COLBERTQABASED\ readers, they are 18\% and 17\%. Conversely, when passages are originated from \XQUADTR, the success rates improve for all readers. Specifically, BM25-based readers achieve a success rate of 49\% for both \WIKITROLD\ and \WIKITRNEW. \DPRBASED\ readers achieve up to 47\% for \WIKITROLD and 44\% for \WIKITRNEW. For \COLBERTQA, these figures reach 48\% when using \WIKITROLD\ and 50\% when using \WIKITRNEW, respectively.\footnote{\REVISIONCOLOR\ Table~\ref{tab:tfidf_ex}, Table~\ref{tab:detailed_info_ex}, and Table~\ref{tab:wordpiece_ex} provide specific examples illustrating the role of Wikipedia passages as distractors.} Although success rates decline for passages retrieved from Wikipedia for all readers, only \COLBERTQA compensates for this drop with a larger increase in success rates for the examples where passages are from \XQUADTR. These numbers explain why reader performance varies as the knowledge source expands. They support our findings from the retriever output that \COLBERTQA is not only a noise-resistant model, but it can also improve its performance as the noise in the knowledge source increases.}

    \begin{table*}
        [ht]
        \centering
        \setlength{\tabcolsep}{3.2pt}
        \resizebox{0.85\textwidth}{!}{%
        \begin{tabular}{l l l l l c c}
            \toprule \textbf{Reader Model}      & \textbf{Retriever Model}        & \textbf{Training Dataset}   & \textbf{Test Dataset}  & \textbf{EM}                   & \textbf{F1}                   \\
            \midrule \REVISIONCOLOR XLM-RoBERTa & BM25 - \emph{Baseline - Sparse} & \SQUADTRTRAINBMYYYYOLD      & \XQUADTRBMYYYYOLD      & \REVISIONCOLOR 37.82          & \REVISIONCOLOR 49.96          \\
                                                &                                 & \SQUADTRTRAINBMYYYYNEW      & \XQUADTRBMYYYYNEW      & \REVISIONCOLOR 37.31          & \REVISIONCOLOR 48.59          \\
            \midrule \REVISIONCOLOR XLM-RoBERTa & DPR - \emph{Baseline - Dense}   & \SQUADTRTRAINBMYYYYOLD      & \XQUADTRBMYYYYOLD      & \REVISIONCOLOR 38.07          & \REVISIONCOLOR 50.40          \\
                                                &                                 & \SQUADTRTRAINBMYYYYNEW      & \XQUADTRBMYYYYNEW      & \REVISIONCOLOR 34.96          & \REVISIONCOLOR 47.36          \\
            \midrule \REVISIONCOLOR XLM-RoBERTa & ColBERT-QA                      & \SQUADTRTRAINCOLBERTYYYYOLD & \XQUADTRCOLBERTYYYYOLD & \REVISIONCOLOR 46.47          & \REVISIONCOLOR 61.22          \\
                                                &                                 & \SQUADTRTRAINCOLBERTYYYYNEW & \XQUADTRCOLBERTYYYYNEW & \REVISIONCOLOR \textbf{47.98} & \REVISIONCOLOR \textbf{61.63} \\
            \bottomrule
        \end{tabular}
        }
        \caption{Reader results for the OpenQA formulation of QA task.}
        \label{tab:openqa_reader_results}
    \end{table*}
    \begin{table*}
        [ht] \resizebox{1\textwidth}{!}{%
        \begin{tabular}{>\REVISIONCOLOR l >\REVISIONCOLOR c >\REVISIONCOLOR c >\REVISIONCOLOR
        c >\REVISIONCOLOR c >\REVISIONCOLOR c >\REVISIONCOLOR c >\REVISIONCOLOR c >\REVISIONCOLOR
        c >\REVISIONCOLOR c >\REVISIONCOLOR c >\REVISIONCOLOR c >\REVISIONCOLOR c}
            \toprule                                                                                                  & \multicolumn{6}{c}{\REVISIONCOLOR \textbf{\WIKITROLD}} & \multicolumn{6}{c}{\REVISIONCOLOR \textbf{\WIKITRNEW}} \\
            \cmidrule(l){2-7} \cmidrule(l){8-13}                                                                      & \multicolumn{3}{c}{\REVISIONCOLOR \textbf{\SQUADTR}}   & \multicolumn{3}{c}{\REVISIONCOLOR \textbf{\WIKITR}}   & \multicolumn{3}{c}{\REVISIONCOLOR \textbf{\SQUADTR}} & \multicolumn{3}{c}{\REVISIONCOLOR \textbf{\WIKITR}} \\
            \cmidrule(l){2-4} \cmidrule(l){5-7} \cmidrule(l){8-10} \cmidrule(l){11-13}

\textbf{Base of Reader Model} & \textbf{Correct}                                       & \textbf{Incorrect}                                    & \textbf{Subtotal}                                    & \textbf{Correct}                                   & \textbf{Incorrect} & \textbf{Subtotal} & \textbf{Correct} & \textbf{Incorrect} & \textbf{Subtotal} & \textbf{Correct} & \textbf{Incorrect} & \textbf{Subtotal} \\
            \midrule

BM25-based - \emph{Baseline - Sparse}                                                           & 432                                                    & 452                                                   & 883                                                  & 19                                                 & 288                & 307               & 424              & 442                & 866               & 20               & 304                & 324               \\
            \midrule

DPR-based - \emph{Baseline - Sparse}                                                            & 441                                                    & 506                                                   & 947                                                  & 12                                                 & 231                & 243               & 409              & 514                & 923               & 7                & 260                & 267               \\
            \midrule

\COLBERTQA-based                                                                                & 544                                                    & 598                                                   & 1142                                                 & 9                                                  & 39                 & 48                & 559              & 560                & 1119              & 12               & 59                 & 71                \\
            \bottomrule
        \end{tabular}
        }
        \caption{Error analysis on the reader outputs on \XQUADTR with respect to
        the source of the passages (\SQUADTR or \WIKITR) and the retriever module
        the reader is based on.}
        \label{tab:openqa_reader_results_error_analysis}
    \end{table*}

    In addition, the OpenQA model for the \COLBERTQA-based reader achieves almost{\REVISIONCOLOR\ 89\%}
    of the standard formulation QA reader results in terms of both EM and F1 scores.
    This result suggests that the OpenQA formulation is productive for low-resource
    and resource-constrained languages, since we can rely on machine-generated noisy
    training data and unstructured knowledge sources.

    \subsubsection{OpenQA Results with Subsampled Test Sets}
    \label{sec:exp:evaluation_with_subsampled_datasets}

\begin{figure*}
        \begin{minipage}{\textwidth}
            \centering
            \includegraphics[scale=.59]{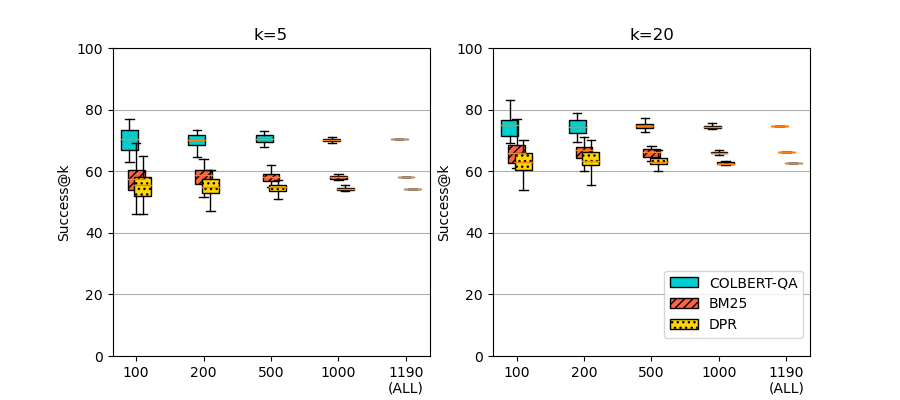}
            \caption{Retriever results for the OpenQA formulation for different sample
            portions of \XQUADTR based on \WIKITRNEW.}
            \label{fig:retriever_results_undersampling}
        \end{minipage}
        \begin{minipage}{\textwidth}
            \centering
            \includegraphics[scale=.59]{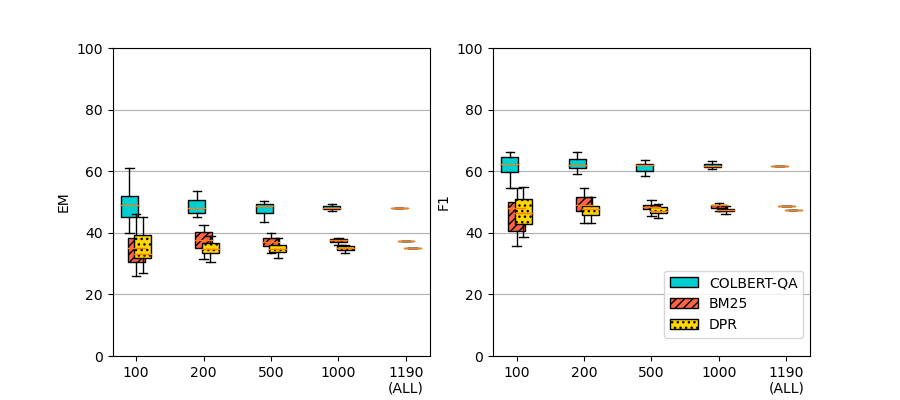}
            \caption{Reader results for the OpenQA formulation for different sample
            portions of \XQUADTR based on \WIKITRNEW.}
            \label{fig:reader_results_undersampling}
        \end{minipage}
    \end{figure*}

    Figure~\ref{fig:retriever_results_undersampling} summarizes the results of the
    experiments {\REVISIONCOLOR\ we conducted using subsampled datasets} for our
    retrievers, and Figure~\ref{fig:reader_results_undersampling} extends the protocols
    to our readers. In each panel, the x-axis tracks the number of assessment examples,
    and the y-axis shows our key metrics.

    Strikingly, with only 100 examples, we can already pretty clearly
    differentiate our BM25-based and DPR-based models from our \COLBERTQA-based
    models. By 200 examples, the systems are dramatically different on all metrics
    for BM25 and \COLBERTQA. As the test sets get larger, the variance of these
    measures gets tighter, as one would expect, but the core conclusions are
    unchanged beyond 200 examples for the models. The results of the BM25 and DPR models
    implicitly suggest that the number of examples needed to differentiate the benchmarked
    models would increase proportional to the competitiveness of the models with respect
    to each other.

    In this setting, we are using the experiments to differentiate three systems,
    but the same logic would apply if we were seeking to determine whether a
    system had truly passed a lower-bound on performance that we set for a
    production system. Overall, these results show that OpenQA systems can be
    evaluated very efficiently. This opens the door to conducting multiple distinct
    evaluations of the same system, which could be crucial for piecing together a
    picture of how the system behaves overall.

    { \REVISIONCOLOR\
 \subsection{Resource Requirements of the Models} \label{sec:exp:resource_usage}

    When aiming to enhance the performance of the models, it is important to also account for their resource consumption, particularly when working with use cases that have limited resources. In this section, we present the resource footprints of the models we utilized in our study.
    The values are based on the model parameters specified in \secref{sec:methods:experimental_protocol}. Varying the model parameters and processing units such as batch size, query and document lengths, and parallel processing or multi-threading can lead to different results.

    \subsubsection{Resource Usage of Retriever Models}\label{sec:openqa_retriever_resource_consumption}

    Table~\ref{tab:retrieval_resource} shows the resource usage of the retriever models for each stage of the retrieval process. We observe that BM25 exhibits the lowest resource footprint across nearly all categories in each phase, offering the fastest indexing and retrieval speeds. In contrast, dense retrievers demonstrate varying resource consumption across different phases.

    \begin{table}[h]\resizebox{1\columnwidth}{!}{%
    \begin{threeparttable}\begin{tabular}{>\REVISIONCOLOR l >\REVISIONCOLOR l >\REVISIONCOLOR l >\REVISIONCOLOR l >\REVISIONCOLOR l}\toprule & \textbf{Resource} & \textbf{BM25} & \textbf{DPR} & \textbf{ColBERT-QA} \\ \midrule \multirow{4}{*}{Training} & \emph{Num of CPU} & - & 1.50 & 1.17 \\ & \emph{Memory} & - & 13.42 GB & 6.14 GB \\ & \emph{Model size} & - & 2.5 GB & 1.3 GB \\ & \emph{Duration} & - & 6 hrs 13 mins & 8 hrs 30 mins \\ \midrule \multirow{4}{*}{Indexing} & \emph{Num of CPU} & 1.27 & 1.07 & 2.76 \\ & \emph{Memory} & 2.57 GB & 19.90 GB & 19.22 GB \\ & \emph{Index size} & 2 GB & 6.5 GB & 114 GB \\ & \emph{Duration} & 14 mins & 1 hr 56 mins & 59 mins \\ & \emph{Speed (D/s)\tnote{*}} & 2610 & 315 & 619 \\ \midrule \multirow{3}{*}{Retrieval/Ranking} & \emph{Num of CPU} & 1.01 & 3.59 & 1.73 \\ & \emph{Memory} & 1.71 GB & 54.41 GB & 78.60 GB \\ & \emph{Duration} & 41 mins & 1 hr 30 mins & 4 hrs 7 mins \\ & \emph{Speed (Q/s)\tnote{**}} & 35.77 & 16.30 & 5.49 \\ \bottomrule\end{tabular} \begin{tablenotes}\item[*] \REVISIONCOLOR D/s: The number of documents indexed per second. \item[**] \REVISIONCOLOR Q/s: The number of questions the retriever can process per second.\end{tablenotes}\end{threeparttable} } \caption{\REVISIONCOLOR Resource consumption of retriever models during training, indexing, and retrieval steps. } \label{tab:retrieval_resource}\end{table}

    In training, DPR uses more memory than \COLBERTQA, mainly because DPR depends on in-batch negatives and its performance is tied to batch size. Additionally, DPR's final model size is twice as large as that of \COLBERTQA because DPR uses two separate BERT models for query and document encoding. In contrast, \COLBERTQA employs a single shared BERT model for both encoders.

    During indexing, BM25 demonstrates exceptional speed when indexing the knowledge source compared to dense retrievers. It also maintains a low memory footprint and minimal index size. DPR and \COLBERTQA have a similar memory footprint. DPR exhibits slower indexing performance relative to \COLBERTQA. (For all models, indexing speed can be boosted using multiple parallel processes.) Additionally, DPR produces a smaller index than ColBERT-QA, in line with the fact that DPR generates fixed-size vectors while ColBERT-QA produces a sequence of token embeddings.

    In the retrieval and ranking step, BM25 maintains the minimum footprint across all categories while achieving the fastest speed. DPR and \COLBERTQA, on the other hand, both demand a considerable amount of memory because they load the entire index into memory during retrieval. The size of memory that DPR and \COLBERTQA consume is in line with the size of their index. Meanwhile, BM25 minimizes memory usage by accessing the index directly from disk.

    \begin{table}[h]\resizebox{0.80\columnwidth}{!}{%
    \begin{threeparttable}\begin{tabular}{>\REVISIONCOLOR l >\REVISIONCOLOR l >\REVISIONCOLOR l >\REVISIONCOLOR l}\toprule & \textbf{Resource} & \textbf{BERT-based} & \textbf{RoBERTa-based} \\ \midrule \multirow{4}{*}{Train} & \textit{Num of CPU} & 0.99 & 1 \\ & \textit{Memory} & 11.05 GB & 14.87 GB \\ & \textit{Model size} & 420M & 1.1 GB \\ & \textit{Duration} & 3 hrs 24 mins & 12 hrs 38 mins \\ \midrule \multirow{3}{*}{Test} & \textit{Num of CPU} & 3.89 & 3.93 \\ & \textit{Memory} & 1 GB & 2.79 GB \\ & \textit{Duration} & 9 mins & 9 mins \\ & \textit{Speed (Q/s)\tnote{*}} & 2.08 & 2.07 \\ \bottomrule\end{tabular} \begin{tablenotes}\item[*] \REVISIONCOLOR Q/s: The number of questions the readers can process per second.\end{tablenotes}\end{threeparttable} } \caption{\REVISIONCOLOR Resource consumption of reader models during training and test steps categorized by the architecture of the model.} \label{tab:reader_resource}\end{table}

    \subsubsection{Resource Usage of Reader Models}\label{sec:openqa_reader_resource_consumption}

    Table~\ref{tab:reader_resource} outlines the resource footprints of the reader models in our study, highlighting their impact and trade-offs across different types of resources. The models were trained using GPUs and then tested exclusively on CPUs to mirror the resource demands of a typical real-world environment.

    The RoBERTa-based model takes approximately three times longer to train compared to BERT-based readers, and its resulting model size is significantly larger than that of BERT-based models. This is largely due to the inherent difference in the number of parameters between the two types of models. Nonetheless, the inference speeds of the models when tested solely on CPUs are fairly similar, though the RoBERTa-based reader attains this speed only by using around three times the memory of BERT-base models. The inference speed also indicates that each CPU core can handle about two questions per second for these readers.

    }

    \section{Discussion}
    \label{sec:discussion} In this section, we discuss the findings of our
    experiments to assess the effectiveness and limitations of the proposed
    approach. We begin by analyzing the results of the standard QA formulation in order
    to observe the potential performance for OpenQA. Then, we delve into the results
    for each component of the OpenQA formulation, namely, the retriever and reader
    modules. Finally, we discuss the required number of gold instances for a reliable
    evaluation of an OpenQA system.

    Perhaps unsurprisingly, {\REVISIONCOLOR\ one of} the best models in the
    standard QA results is the one that begins with BERTurk parameters and trains
    on our in-language dataset. However, using multilingual embeddings (mBERT) is
    competitive with this, and the only large gap in performance is between these two
    models and the mBERT model trained on English data. This aligns with other
    recent findings and shows that in-language training data is the real
    differentiator, even if it is potentially noisy MT data.
    {\REVISIONCOLOR\ Furthermore, \XLMROBERTA outperforms BERTurk, highlighting the impact architectural improvements can have on performance. These results also point out the potential for developing in-language variants of RoBERTa, which can outperform \XLMROBERTA, as observed in other languages~\cite{scheible2020gottbert, martin-etal-2020-camembert, le-etal-2020-flaubert-unsupervised}.}

    The success of OpenQA systems highly depends on the performance of their retriever
    models. The baseline BM25 and DPR retrievers were able to capture at most, respectively,
    {\REVISIONCOLOR\ 56.30\% and 52.10\%} of the target passages, whereas the ColBERT-QA
    retriever increased this to {\REVISIONCOLOR\ 77.05\%}. As such, the ColBERT-QA
    retriever model increased the likelihood of the reader model finding more relevant
    passages to answer questions. These results also indicate that there is a room
    for further improvement within the retrieval module
    {\REVISIONCOLOR\ to find more passages containing relevant answer spans for the questions}.

    { \REVISIONCOLOR\
Furthermore, the loss of some answer span locations within the target paragraphs after translation does not necessarily hinder the performance of the retriever models; in fact, weak supervision presents an advantage in such cases. This is because retrievers can instead fetch other suitable paragraphs from the knowledge source, resulting in noise-resistant retriever results. Consequently, this increases the number of examples with positive passages in training bootstrap datasets for the retriever models and training/test datasets for the reader models. The noise-robustness of retrievers in the weak supervision setting also means that we can rely on datasets of question--answer pairs, with no need to include or annotate passages. The superior performance of DPR- and \COLBERTQA-based reader models over the BM25-based reader model, particularly when employing \WIKITROLD, underscores how much weak supervision facilitates retriever models in benefiting from this augmentation more effectively. Notably, the substantial and consistent improvement in the \COLBERTQA-based reader model is evident both with \WIKITROLD and \WIKITRNEW, and it indicates the superior capacity of \COLBERTQA to derive benefits from weak supervision compared to DPR. }

    {\REVISIONCOLOR\
Additionally, we note that expanding the knowledge source can potentially impact the performance of the OpenQA system in two ways. It may either enhance or degrade the performance of the OpenQA system, depending on the capacity of the retriever model to effectively navigate the increased noise in the expanded knowledge source to identify more positive passages. We observe that \COLBERTQA is more capable of navigating this challenge compared to BM25 and DPR models. Consequently, as the knowledge source gradually expands over time, \COLBERTQA leads to a significant enhancement in the overall success of its reader towards its upper limit capped by the standard QA reader results. }

    In the standard QA formulation, the reader models achieved a maximum EM score of
    {\REVISIONCOLOR\ 52.18\%}, which sets the upper bound for the OpenQA reader
    models. The ColBERT-QA based OpenQA reader results demonstrated that we can preserve
    almost {\REVISIONCOLOR\ 89\%} of this score without requiring gold passages as
    input. This relaxation of the input requirement provides a significant
    advantage for developing cost-effective QA systems in low-resource
    {\REVISIONCOLOR\ language contexts}.

    Although we do not require input gold passages at training time, we still need
    a labeled dataset at test time. Our experiments revealed that a few hundred
    evaluation examples may be enough to confidently differentiate the performance
    of models. {\REVISIONCOLOR\ Lowering the requirement for the number of gold examples is highly important, especially for low-resource language scenarios where the necessary human resources are not widely available, e.g. endangered Indigenous Languages~\cite{ebrahimi-etal-2023-findings, shode-etal-2023-nollysenti}. In other language scenarios, t}his
    result not only helps limit the cost of obtaining QA systems but also paves
    the way for obtaining multiple evaluation datasets with different
    characteristics to better reflect the overall picture of the models in
    production.

    In summary, our proposed system does not require gold datasets during training,
    but instead utilizes existing unstructured knowledge sources and MT systems to
    create a machine-translated labeled training dataset
    {\REVISIONCOLOR\ for an OpenQA application in Turkish as a use case for low-resource language contexts}.
    The potential noise in the resulting training dataset can be overcome with the
    help of the weak supervision used in the OpenQA formulation, which is
    resilient to noisy data. As a result, we have shown that a cost-effective QA
    system is feasible for low-resource {\REVISIONCOLOR\ language contexts} when
    we shift our focus to the OpenQA formulation. { \REVISIONCOLOR\


    \section{Adaptation and Generalization: Extending Our Method to Diverse Language Contexts}\label{sec:generalization}

    \begin{table*}[tp] \noindent\makebox[\linewidth]{\rule{\linewidth}{0.2pt}} \begin{enumerate}\item \label{step:mt} \textbf{Building a QA dataset in a new language context.} Create an extractive reading comprehension dataset in the target language context as follows: \begin{itemize}\item \textbf{Training dataset:} Translate an existing extractive QA dataset using automatic translation.

    \item \textbf{Evaluation dataset:} Obtain an evaluation dataset containing as few as 200 question--answer pairs.\end{itemize}

    \item \label{step:knowledge_source} \textbf{Compiling a knowledge source.} Create a knowledge source by compiling passages answering the questions in the extractive QA dataset obtained in the previous step.

    \item \textbf{Training an OpenQA retriever.} Train a neural retriever weakly supervised by the training{\REVISIONCOLOR\ bootstrap dataset derived from the training dataset and knowledge source prepared in the previous steps.}

    \item \textbf{Creating an OpenQA reader.} Train an off-the-shelf reader using the extractive QA training dataset, where the contexts are now provided by the OpenQA retriever.\end{enumerate} \noindent\makebox[\linewidth]{\rule{\linewidth}{0.2pt}} \caption{Our general method for creating OpenQA systems in low-resource languages efficiently.} \label{tab:overall_method}\end{table*}

    We presented a general purpose method for training OpenQA in \textit{low resource contexts} and use Turkish QA as a case study for showing its efficacy. In this section, we detail the steps for adapting our method to different low-resource contexts other than the presented approach for Turkish. A summary of our overall method is shared in Table \ref{tab:overall_method}.

    \paragraph{Step 1. Building a QA dataset in a new language context} The main blocker for building OpenQA systems in low-resource contexts is the lack of a QA dataset itself. In our case study, we resolve this issue for Turkish by translating a readily available extractive QA dataset in English to Turkish using an automatic translation service. Similar strategies can be applied to different QA datasets for other languages given the availability of the translation tools between the original and the target languages.

    When obtaining a QA dataset through machine translation, one can translate the contexts, questions, and answer spans contained in a typical extractive dataset to the target language. However, the standalone translation of some of the answer spans may be different from their corresponding translations in the context paragraphs, due to the contextual nature of translation. In our case study with Turkish, we employ approximate string matching as a post-processing step to recover some of these answer spans, taking into account the morphological complexity of the Turkish language. When adapting our method to a different target language, one can devise different post-processing methods considering the linguistic characteristics of the target language. For example, a basic stemming and lemmatization algorithm could also be used for languages that have fewer inflectional forms than Turkish. Similarly, accent normalization could also be helpful as a post-processing step for languages with accent-rich alphabets, like Vietnamese~\cite{nguyen2016text, le-etal-2022-vimqa}.

    At the end of this process, we successfully obtained a QA dataset in the target language, which we used as our training data. In our case study, this training dataset included 81K examples, for which 61K had their answer spans recovered. The size of our training dataset can be used as a reference when creating training datasets for other low-resource contexts, whether or not translation tools are utilized.

    Although we can bootstrap our training dataset using a translation service, we need a high quality evaluation dataset for reliable assessment. In our method, we utilize an external human-annotated dataset, \XQUAD~\citep{artetxe-etal-2020-cross}, that supports 10 other languages. Similar datasets can also be employed for different language~(e.g. XQA~\cite{liu-etal-2019-xqa}; MLQA~\cite{lewis-etal-2020-mlqa}; MKQA~\cite{longpre-etal-2021-mkqa}) and domain scenarios (e.g. e-commerce~\cite{shen-etal-2023-xpqa}, medical~\cite{pergola2021boosting,daniel-etal-2019-towards}, legal~\cite{ghosh-etal-2023-dale}, finance~\cite{sun:kdd23ws}, customer service~\cite{zheng-etal-2023-dialogqae}, space~\cite{darm-etal-2023-discosqa}). Nevertheless, the availability of such datasets remains limited and may lack support for the targeted low-resource context. In such scenarios, our analysis showcased in \secref{sec:exp:evaluation_with_subsampled_datasets} suggests that a dataset consisting of merely 200 examples can still provide valuable insights.

    \paragraph{Step 2. Compiling a knowledge source} We then create a knowledge source that will serve as the corpus for OpenQA in our target context. We used Wikipedia as our knowledge source but other knowledge sources can also be used depending on the use case and the availability of resources in the target low resource context.

    One specific limitation we encountered when using Turkish Wikipedia was that the majority of target paragraphs from our QA datasets were missing in Turkish Wikipedia. To tackle this issue, we augmented our knowledge source by adding the paragraphs from our QA dataset. This allowed us to create an OpenQA model even under limited resources. Such an approach can be applied to other scenarios with comparable limitations.

    In order to optimize the performance of the retriever models, we segment the passages in the resulting knowledge source into smaller chunks, ensuring that the number of tokens per chunk fits into the context of the retriever models. While 100 words proved sufficient for English, we found it necessary to reduce this to 75 words for Turkish due to the tokenizer for Turkish generating longer token sequences. Likewise, the chunk size can be adjusted based on the linguistic characteristics of the target language and the requirements of the chosen retriever.

    \paragraph{Step 3. Training an OpenQA retriever}

    We use BM25 as a lightweight retriever to prepare the bootstrap training datasets for the advanced retrievers. We customize the BM25 retriever by incorporating a morphological stemmer tailored for Turkish. The default or custom analyzers for other languages can be employed for similar adaptations~\cite{clavie2023jacolbert}. When training the advanced retrievers in our OpenQA system, we used BERTurk~\cite{stefan_schweter_2020_3770924} to customize the tokenizers of our chosen retrievers and for weight initialization, thereby customizing them for Turkish. Likewise, for other languages, comparable in-language BERT variants can be utilized for adaptation. In cases where an in-language BERT model is not accessible for the desired target language, the multilingual BERT model~\citep{artetxe-etal-2020-cross} offers a viable alternative.

    \paragraph{Step 4. Creating an OpenQA reader} The last step of our method is training an OpenQA reader using our retriever along with our extractive QA training dataset. To do so, we take XLM-RoBERTa~\cite{conneau-etal-2020-unsupervised} as an off-the-shelf model in our target language, which showed superior performance in our standard QA experiments. Other models, such as in-language variants of BERT~\cite{devlin-etal-2019-bert}, ALBERT~\cite{lan2020albert} and RoBERTa~\cite{liu2019roberta}, can be also selected depending on the availability of resources and needs of the target language context. We fine-tune the reader in an extractive QA setting using the training dataset obtained from the output of the retrievers employed in our experiments. }
    \section{Conclusion}
    \label{sec:conclusion}

    In this paper, we obtained an affirmative answer to our core research question,
    \emph{Can we develop cost-effective OpenQA systems for low-resource{\REVISIONCOLOR\ language contexts}
    without requiring a gold training dataset?} We further expanded our question to
    explore the minimum test set sizes required to reliably evaluate the performance
    of OpenQA models. Our findings help pave the way to transferring the rapid
    advancements made in English QA to non-English QA systems. Moreover, this new
    avenue not only allows one-way transfer of advancements, but also establishes a
    virtuous cycle between English OpenQA and non-English OpenQA systems,
    promoting mutual progress.{\REVISIONCOLOR\ Furthermore, the proposed methodology can also benefit certain domain-specific scenarios, even in high-resource languages like English, where data remains scarce.}

    We presented a general method for creating efficient and effective OpenQA systems
    for low-resource{\REVISIONCOLOR\ language contexts}, and we illustrated the method
    with a case study of Turkish. Our overall method is summarized in Table~\ref{tab:overall_method}
    as a recipe. As part of this, we introduced \SQUADTR, a Turkish QA dataset derived
    by automatically translating SQuAD2.0~\citep{rajpurkar-etal-2018-know}. We
    showed that \SQUADTR can straightforwardly be used to train high quality
    OpenQA systems and benchmark different types of models, and we supported this assessment
    with detailed qualitative analysis. In addition, we provided evidence that the
    success of the OpenQA system is notably enhanced by expanding the knowledge source
    {\REVISIONCOLOR\ depending on the retriever's capability on navigating the potential increase in noise accompanying the knowledge source expansion},
    and we showed that a relatively small number of gold test cases may be
    sufficient to obtain confident assessments of the quality of these systems.

    The key to creating these systems in non-English languages so efficiently is
    the move from standard QA to OpenQA. In doing this, we greatly simplify the
    process of creating gold examples, which has been a barrier for the advancement
    in QA systems for low-resource languages. In OpenQA, these datasets are just question--answer
    pairs, completely eliminating the necessity for answer span annotation.
    Consequently, these datasets can now be acquired through automatic translation
    from the abundant resources available in English. The OpenQA task is also
    arguably more \textit{relevant}, in that it comes much closer than standard QA
    to simulating the experience of searching a real-world knowledge store like
    the Web. Thus, we hope not only to have removed obstacles to creating QA
    systems for low-resource languages like Turkish, but we also hope to have helped
    motivate the OpenQA task more generally, as a step towards QA systems that can
    truly meet the information needs of real-world users. We publicly share our code,
    models, and data to encourage future research.


    \printcredits

    \section*{Acknowledgments}
    This research was supported by the AWS Cloud Credits for Research Program (formerly
    AWS Research Grants)\footnote{Disclaimer: The AWS Cloud Credits for Research Grant
    was awarded before the corresponding author joined Amazon.} and the Turkish
    Directorate of Strategy and Budget under the TAM Project number 2007K12-873. E.~Budur
    is thankful for the support provided by Council of Higher Education (Y\"{O}K)
    100/2000 Graduate Research Scholarship Program.

    The authors gratefully acknowledge that the computational parts of this study
    have been mostly performed at Bo\u{g}azi\c{c}i \mbox{TETAM DGX-1} GPU Cluster
    and partially carried out at T\"{U}B\.{I}TAK ULAKB\.{I}M High Performance and Grid
    Computing Center (TRUBA resources)\footnote{https://www.truba.gov.tr} and
    Stanford NLP Clusters.

    We thank Alara Dirik, Almira Ba\u{g}lar, Berfu B\"{u}y\"{u}k\"{o}z, Berna Erden,
    G\"{o}k\c{c}e Uludo\u{g}an, Havva Y\"{u}ksel, Melih Barsbey, Murat Karademir,
    Selen Parlar, Tu\u{g}\c{c}e Ulutu\u{g}, Utku Yavuz for their support on our application
    for AWS Cloud Credits for Research Program and Fatih Mehmet G\"{u}ler and
    Alican Acar for the valuable advice, discussion, and insightful comments.

    \printassistiveaitechdisclosure

    \bibliography{bibliographies/anthology, bibliographies/manual_entries}
    \bibliographystyle{model1-num-names}
\end{document}